\documentclass[sigconf, 10pt, nonacm=true, balance=false]{acmart}

\usepackage{listings}
\usepackage{color}
\usepackage{float}

\definecolor{codegreen}{rgb}{0,0.6,0}
\definecolor{codegray}{rgb}{0,0,0}
\definecolor{codepurple}{rgb}{0.58,0,0.82}
\definecolor{backcolour}{rgb}{0.95,0.95,0.92}
 
\lstdefinestyle{mystyle}{
    backgroundcolor=\color{backcolour},   
    commentstyle=\color{codegreen},
    keywordstyle=\color{magenta},
    numberstyle=\tiny\color{codegray},
    stringstyle=\color{codepurple},
    basicstyle=\fontsize{8}{10}\ttfamily,
    breakatwhitespace=false,         
    breaklines=true,                 
    captionpos=b,                    
    keepspaces=true,                 
    numbers=left,                    
    numbersep=5pt,                  
    showspaces=false,                
    showstringspaces=false,
    showtabs=false,                  
    tabsize=2,
    frame=single
}
 
\def\BibTeX{{\rm B\kern-.05em{\sc i\kern-.025em b}\kern-.08emT\kern-.1667em\lower.7ex\hbox{E}\kern-.125emX}}

\begin{document}

%
\title{Horus: Using Sensor Fusion to Combine Infrastructure and On-board Sensing to Improve Autonomous Vehicle Safety }

%
\author{Sanjay Seshan}
\affiliation{%
  \institution{Fox Chapel Area High School, Grade 10}
  \city{Fox Chapel}
  \state{PA}
}

%
\renewcommand{\shortauthors}{Seshan}

%

\begin{abstract}
Studies predict that demand for autonomous vehicles will increase tenfold between 2019 and 2026.  However, recent high-profile accidents have significantly impacted consumer confidence in this technology. The cause for many of these accidents can be traced back to the inability of these vehicles to correctly sense the impending danger. In response, manufacturers have been improving the already extensive on-vehicle sensor packages to ensure that the system always has access to the data necessary to ensure safe navigation. However, these sensor packages only provide a view from the vehicle's perspective and, as a result, autonomous vehicles still require frequent human intervention to ensure safety. 

To address this issue, I developed a system, called Horus, that combines on-vehicle and infrastructure- based sensors to provide a more complete view of the environment, including areas not visible from the vehicle. I built a small-scale experimental testbed as a proof of concept. My measurements of the impact of sensor failures showed that even short outages (\textasciitilde1 second) at slow speeds (\textasciitilde25 km/hr scaled velocity) prevents vehicles that rely on on-vehicle sensors from navigating properly. My experiments also showed that Horus dramatically improves driving safety and that the sensor fusion algorithm selected plays a significant role in the quality of the navigation.  With just a pair of infrastructure sensors, Horus could tolerate sensors that fail 40\% of the time and still navigate safely. These results are a promising first step towards safer autonomous vehicles.
\end{abstract}

%
\keywords{robotics, autonomous vehicles, engineering, vision systems, sensor fusion}

%
\maketitle

\section{Introduction}
\label{sec:intro}

Market research~\cite{2018.allied} predicts that the demand for autonomous vehicles will increase tenfold between 2019 and 2026.  However, recent high-profile incidents, such as the fatal accidents with Uber~\cite{2018.nytimes} and Tesla~\cite{2018.wired} self-driving vehicles, have significantly impacted consumer confidence in this technology. A recent study by AAA~\cite{2018.aaa} reports that three-quarters of Americans are too afraid to ride in an autonomous vehicle and two-thirds feel less safe when self-driving cars are present. In order for autonomous cars to reach their potential, they must be made safer and more reliable. 

Existing approaches to improving autonomous driving focus on extensive sensor packages on-vehicle to ensure that the system always has access to the data necessary to ensure safe navigation. For example, CMU's Boss, which was used in the DARPA Urban Challenge, contains 4 radars, 8 LIDAR and 1 submeter-accurate GPS sensor. While this was an early prototype, modern implementations use similar sensor suites. Tesla Autopilot 2.0's hardware consists of 8 cameras, 1 radar and 12 ultrasonic sensors~\cite{2017.Tesla} and Uber's most recent self-driving car uses 7 Cameras, 1 LIDAR, and multiple radar and ultrasonic sensors~\cite{2016.Etherington}. 

Even with all these sensors, current autonomous cars too often still require driver intervention to ensure safety. Tesla's current systems are considered SAE Level 2 (Partial Automation)~\cite{2017.SAE} and Uber's vehicles have required human intervention more than once per mile traveled. Alphabet/Waymo has reported far fewer interventions; however it is not clear what driving conditions were used at the time~\cite{2017.Bhuiyan}.

The details of these interventions are often relatively vague; for example, Waymo's intervention causes included ``software discrepancy'', ``weather conditions during testing'', ``recklessly behaving road user'', ``unwanted maneuver of the vehicle'', ``perception discrepancy'', ``incorrect behavior prediction of other traffic participants'', ``construction zone during testing'', ``emergency vehicle during testing'' and ``debris in the roadway''. There is a bit more information available about CMU Boss, which documented several specific interventions: 1) A dust cloud kicked up by the car was interpreted as an obstacle; 2) The Boss team incorrectly defined the shape of the traffic lanes causing the car to be overly cautious; 3) A pair of parked cars  were partially in the road and Boss's system incorrectly concluded that it could not navigate around the cars while staying within its lane; 4) An oncoming vehicle was partially in Boss's lane and occluded Boss's view. This caused Boss's sensing system to estimate another vehicle's orientation incorrectly. These failures were primarily a result of sensing or configuration failures~\cite{2008.Urmson}. 

\begin{figure}[t]
  \centering
  \includegraphics[width=\linewidth]{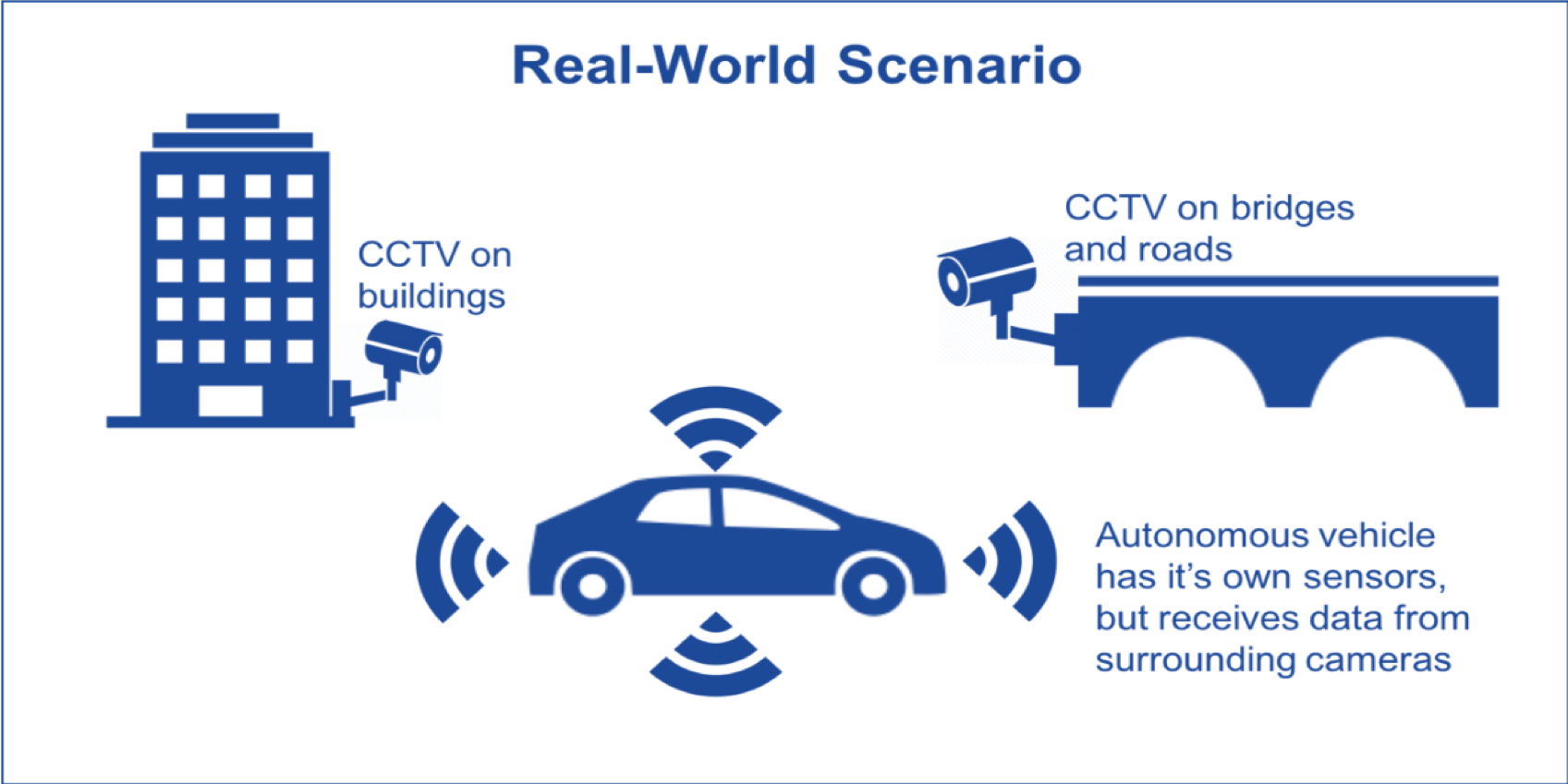}
  \caption{Real-world application}
  \Description{Diagram of scenario.}
  \label{fig:real_world}
\end{figure}

What if the autonomous vehicles could use the data from the sensors that already exist in the surrounding infrastructure such as cameras mounted on roads and bridges (Figure~\ref{fig:real_world})? These sensors often provide a wider view of a car's surroundings than the on-vehicle sensors. Access to current and historical infrastructure sensor observations could have helped address many of the issues reported with the Waymo and CMU Boss systems. For example, historic readings could have helped identify locations with common problematic readings (e.g. dust cloud and misconfigured lane sizing) and additional sensor readings could have addressed issues with accuracy (e.g. position of parked cars) and coverage (e.g. views into blocked areas).

While the vehicle would still need to operate correctly without these infrastructure sensor readings, I believe that their use could lead to improvements in safety, lower the frequency of driver interventions, and an overall improvement in driving quality (e.g. fewer sudden adjustments or higher overall speed).
This research study explores the possibility of using data from these external sensors to enable navigation and provide anti-collision data. The hypothesis being investigated is that an autonomous vehicle can navigate solely using data from external sensors. The research will be conducted in multiple phases.

Phase I of this project explored the challenges associated with having a vehicle navigate solely with the use of a single infrastructure sensor. A key concern in using infrastructure sensors is that observations must be transmitted over a wireless network to the vehicle. If the network is slow or unreliable, the vehicle may not get observations in time to make driving decisions. I examined how network reliability affects the quality of autonomous vehicle navigation. The experiments successfully showed that infrastructure control of an autonomous vehicle is resilient to both packet loss and delay when the UDP protocol is used to transfer data. 

In Phase II of this project, I explore the challenges of combining the observations from multiple infrastructure cameras as well as an on-vehicle camera. A key concern for this phase is reconciling data from multiple sources. Each camera has a partial or obstructed view of the world. The data needs to be stitched together to create reliable autonomous vehicle navigation. The hypothesis is that fusing data from multiple sensors offers more reliable navigation than just using an on-vehicle camera. 

To do this study, I create a new prototype autonomous vehicle and add additional sensors to both the vehicle and environment (Section~\ref{sec:materials}). I design a system called Horus (Section~\ref{sec:software}) that fuses the data from multiple on-vehicle and infrastructure cameras. My experimental results (Section~\ref{sec:results}) shows the feasibility of this approach. My results indicate that even with just a pair of infrastructure sensors to help guide, a vehicle can make autonomous driving robust even when sensors fail to get useful readings 40\% of the time. 
\section{Materials \& Prototype Design}
\label{sec:materials}

\begin{figure}[t]
  \centering
  \includegraphics[height=4cm,keepaspectratio]{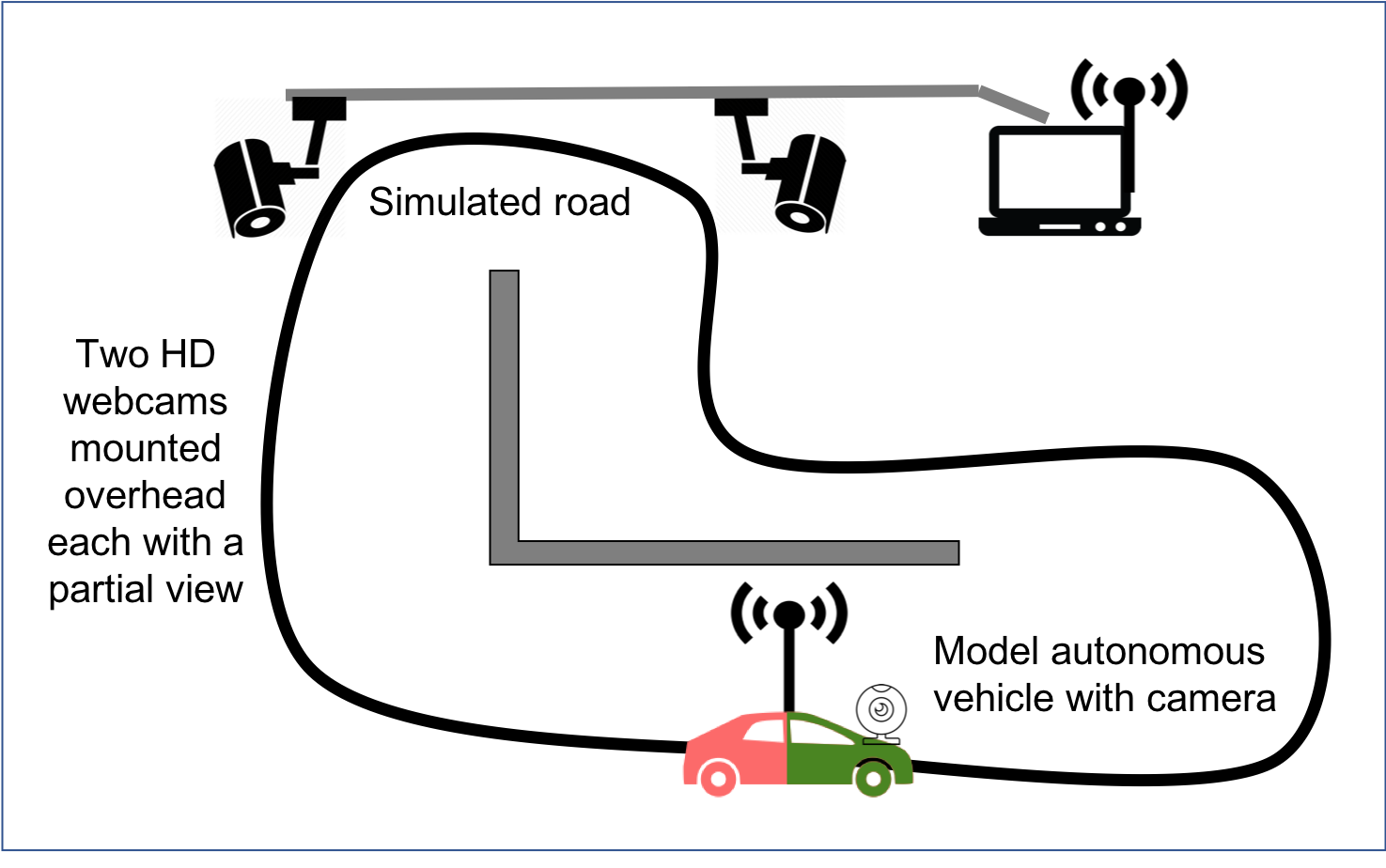}
  \caption{Image of Phase II Experimental Testbed.}
  \label{fig:exp_setup2}
\end{figure}

Phase II of this project focuses on the effective use of multiple on-vehicle and in-infrastructure sensors. Key issues I explore include determining how to combine observations from different sensors that observe the vehicle at the same time, how to use sensors that only provide partial coverage of the vehicle's route, and how to use both on-vehicle and infrastructure sensors to guide the vehicle. To support experiments about these issues, the experimental testbed must model a more complex simulated road environment that limits sensor views and a simulated infrastructure with multiple sensors. Based on these needs, my experimental setup for Phase II (shown in Figure~\ref{fig:exp_setup2}) uses the following materials:

\begin{itemize}
    \item A $2m \times 2m$ white board with a black line to represent the road. The area is walled to achieve blocked views

    \item A Linux Intel i7-based PC with two HD-quality webcams to simulate the infrastructure sensors and computation resource.
\end{itemize}

\begin{figure}[t]
  \centering
  \includegraphics[width=\linewidth,keepaspectratio]{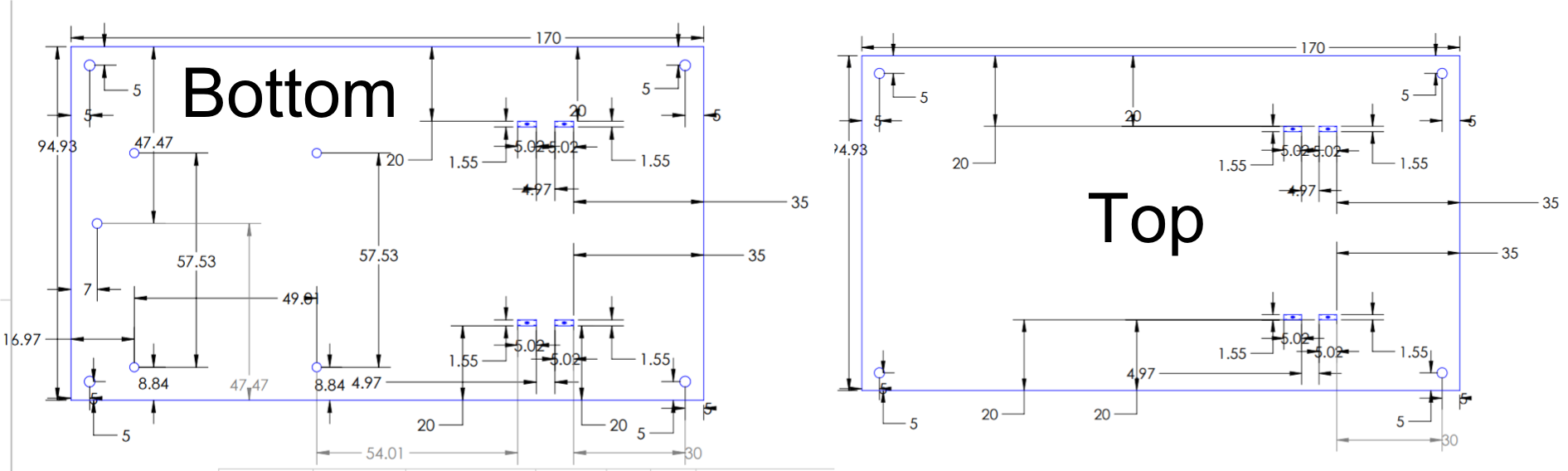}
    \caption{CAD drawings of vehicle mounting plates.}
  \label{fig:robot_draw}
\end{figure}

\begin{figure}[t]
  \centering
  \includegraphics[width=\linewidth,keepaspectratio]{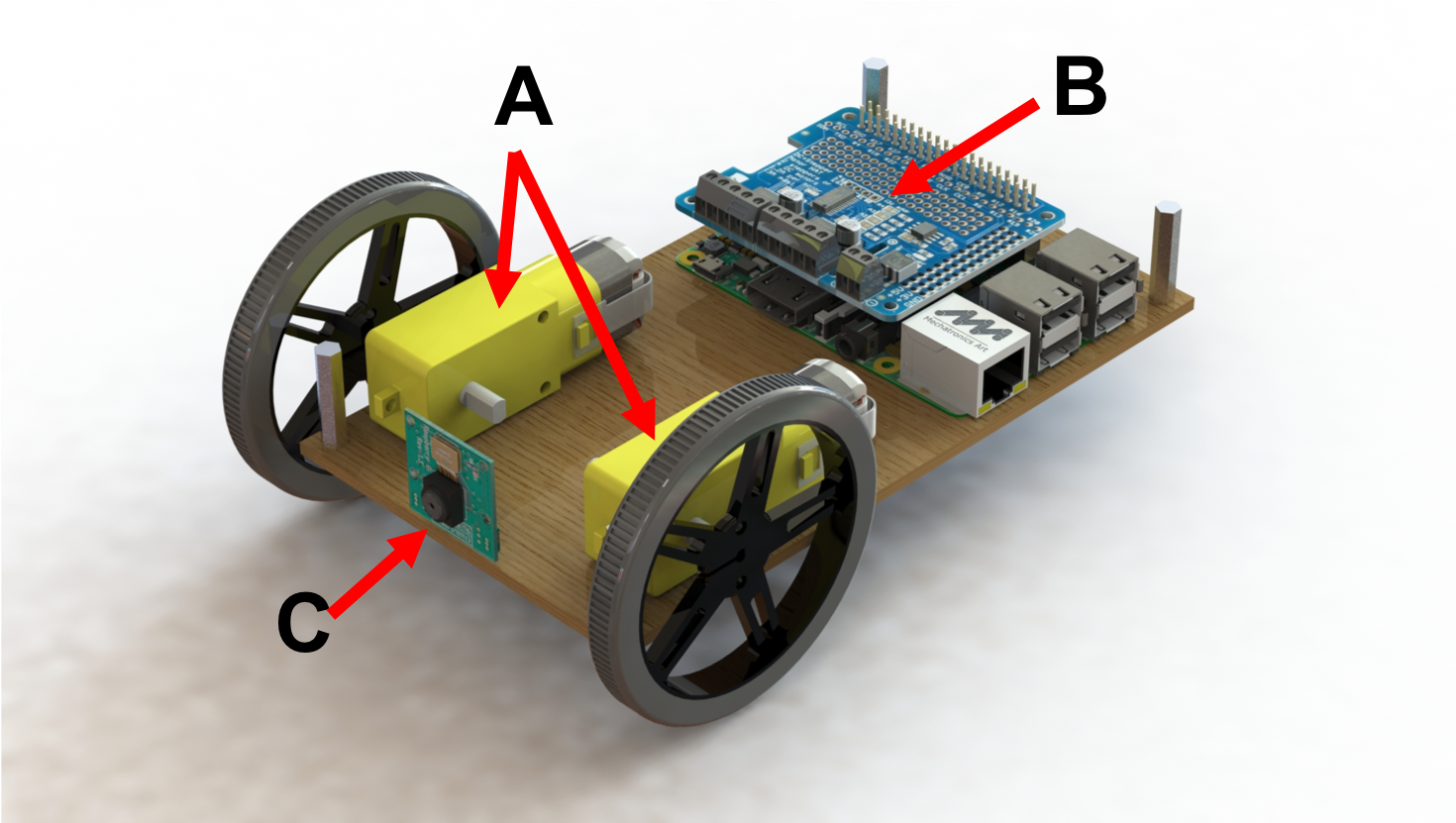}
  \caption{Rendering of SolidWorks design of vehicle bottom mounting plate and hardware. Labels identify (A) DC motors, (B) Raspberry PI and DC motor controller HAT and (C) Raspberry Pi camera.}
  \label{fig:robot_bottom}
\end{figure}

\begin{figure}[t]
  \centering
  \includegraphics[width=\linewidth,keepaspectratio]{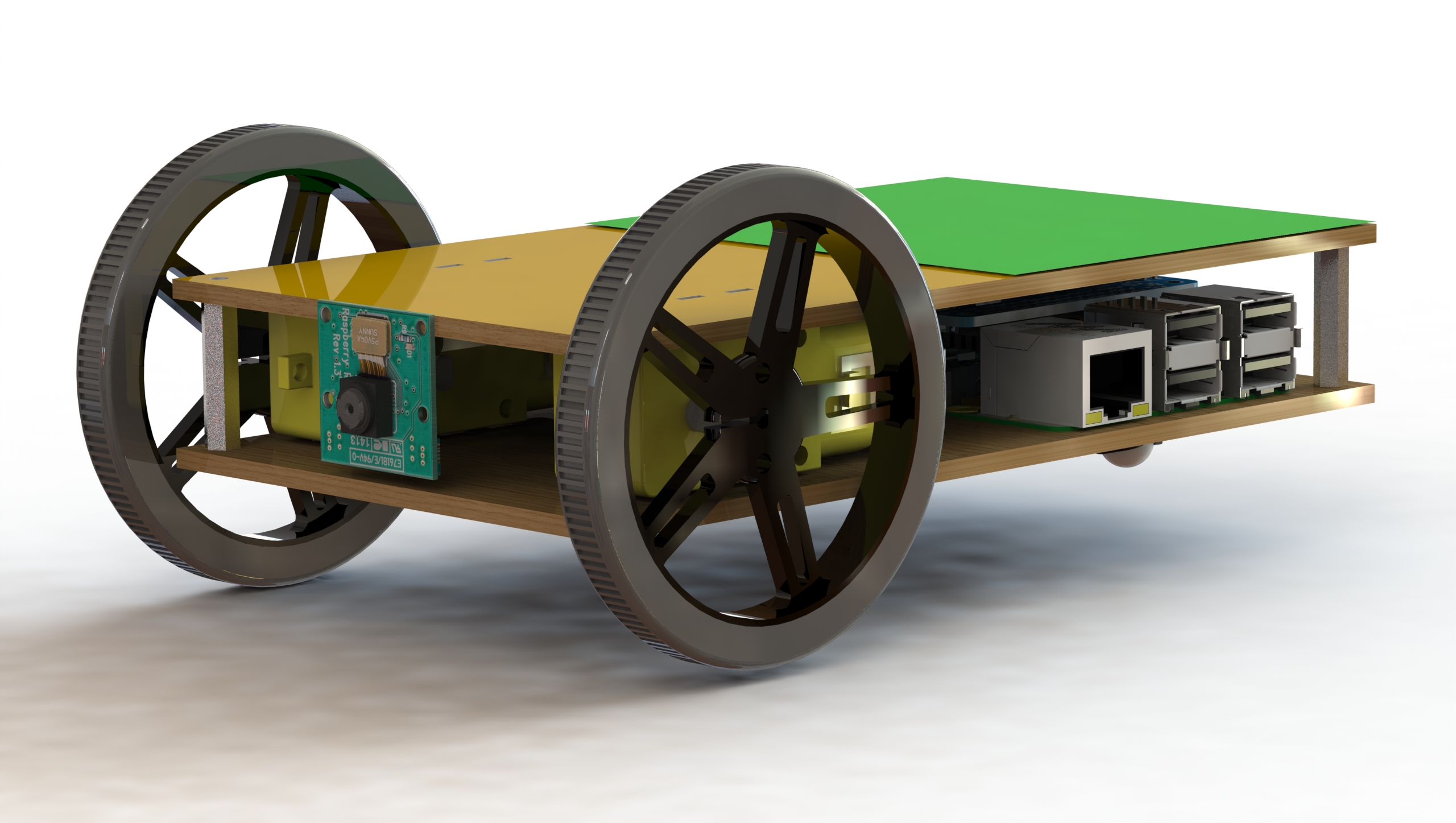}
  \caption{Rendering of SolidWorks design of full vehicle.}
  \label{fig:robot_full}
\end{figure}

In addition, Phase II needs a vehicle that can support video capture and image processing. To meet this need, I create a custom vehicle design using:

\begin{itemize}
    \item A Raspberry Pi 3 B+ MicroComputer~\cite{pi} running Linux.
    
    \item Two Adafruit DC Gearbox Motors~\cite{motor} and wheels.

    \item An Adafruit DC motor controller HAT for the Raspberry Pi~\cite{motorhat}.
    
    \item A Raspberry Pi Camera module V2~\cite{picamera}. 
    \item Battery pack to power Pi and motors.
\end{itemize}

I designed the physical structure of the vehicle using SolidWorks 3D CAD Design Software~\cite{solidworks} and laser-cut my designs from wood using a GlowForge laser cutter~\cite{glowforge}. The design of the two mounting plates that form the primary vehicle structure is shown in Figure~\ref{fig:robot_draw}. The dimensions of the plates were chosen to be representative of the aspect ratio of a typical car. A rendering of the bottom mounting plate with motors, Raspberry Pi and camera attached is shown in Figure~\ref{fig:robot_bottom}. The motors are directly connected to the wheels to avoid the need for gearing and associated movement inaccuracies. The vehicle has separate left and right motors and steers by giving different power to the two motors. A small ball-bearing skid was added to the rear to allow the vehicle to use this form of skid-steer based drive. While car steering is done differently, this design is simpler and the differences between the two steering mechanisms do not impact the investigation of multi-sensor issues. The Pi camera was placed in a location where it could get an unobstructed view of the road for the purpose of autonomous navigation. Finally, to make the vehicle easier to track using cameras, my design for the top plate for the vehicle could easily accommodate the addition of a color panel.  Figure~\ref{fig:robot_full} shows a rendering of the full robot with a green and orange color panels added to the top plate. My testbed vehicle is approximately 1:30th scale of a real vehicle and my experiments run the vehicle at a scaled speed of 25km/hr (or a real speed of 0.25 m/s). 
\section{Software Design}
\label{sec:software}

In this section, I provide an overview of the Horus system software; some of the error checking and details have been removed from the code segments below to improve readability. I have made the entire code for the project available in the Appendix~\ref{sec:code} of this document and online~\cite{2019.code}. 

The software for the project is written in three parts. The first part (Section~\ref{sec:on-vehicle}) of the software runs on the on-vehicle Raspberry Pi and analyzes the video from the on-vehicle camera using the OpenCV~\cite{2018.opencv} computer vision processing library. This video processing code determines the location of the road in the frame and uses the location estimate along with Proportional-Integral-Derivative (PID) control system to compute the steering needed to center the vehicle on the road. Second, (Section~\ref{sec:infrastructure}) code on a PC running the Linux operating system collects video from an infrastructure camera and similarly analyzes it to identify the relative location of the vehicle to the road. This code also computes a steering correction using a PID control system. The third component (Section~\ref{sec:vcs}) of the software runs on the Raspberry Pi and receives the steering instructions from different camera systems (both infrastructure and on-vehicle). It combines these instructions in a weighted fashion and applies the output to control the wheel motors. In the remainder of this section, I describe each of these components in detail and how they combine to create the Horus system. 

\subsection{On-Vehicle Sensor Processing}
\label{sec:on-vehicle}

\begin{figure}[t]
  \centering
  \includegraphics[width=\linewidth]{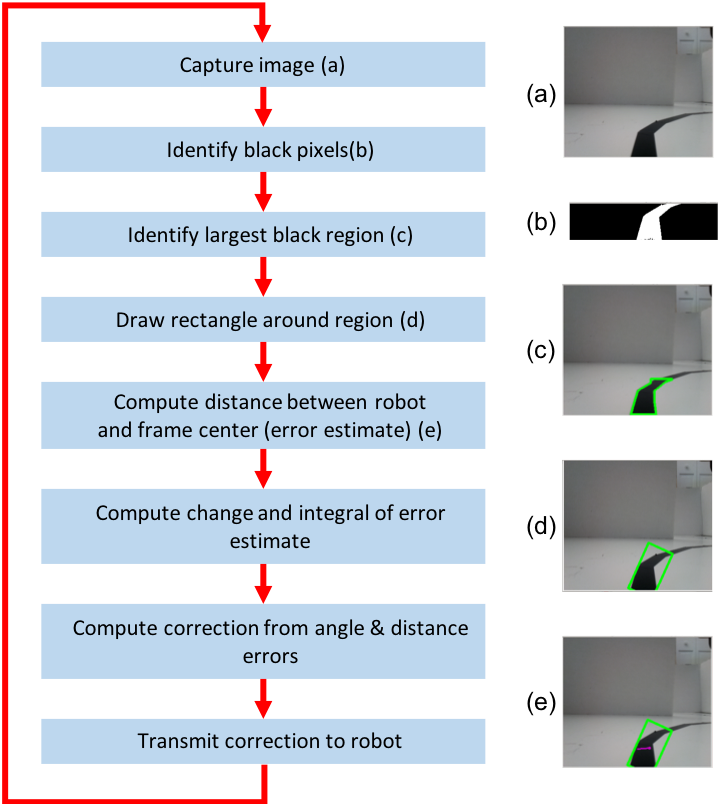}
  \caption{A flowchart of the on-vehicle sensor processing code.}
  \label{fig:code_flowchart}
\end{figure}

The first step in the on-vehicle processing is to locate the road in the video feed and determine the vehicle's location and direction relative to the road. A flow chart of this code is shown in Figure~\ref{fig:code_flowchart}. I describe each of steps in detail below. 

The first step in the infrastructure software is to capture a video frame. The output of the first step is shown in Figure~\ref{fig:code_flowchart}(a). 
While the Pi camera is capable of capturing at high quality ($1920\times1080$ resolution), I intentionally use a lower camera resolution ($320\times240$) to speed up the subsequent processing steps. This enables the system to process frames at a higher rate and give more frequent navigation feedback to the vehicle. I also convert the image representation to use HSV (hue, saturation and value) instead of RGB (red, green, blue) for later processing steps. This is done with the below code:

\begin{lstlisting}[language=Python]
rawCapture = PiRGBArray(camera, size=(320, 240))
for frame in camera.capture_continuous(rawCapture, format="bgr", use_video_port=True):
    full_img = frame.array
    imgHSV = cv2.cvtColor(full_img,cv2.COLOR_BGR2HSV)
\end{lstlisting}

The next steps identify the road in the video frame. The first part of this process is to identify just the black pixels within the image. 
This is done using the function \texttt{FindColor}, shown in Figure~\ref{fig:code_colors}. The range of colors that the function searches for are described using the \texttt{lower\_col} and \texttt{upper\_col} parameters. Line 4 in the code uses openCV to identify all pixels between these two colors in the HSV color space. The result, shown in Figure~\ref{fig:code_flowchart}(b), can be a relatively noisy image since there are often stray pixels that look black. To eliminate this noise, Lines 6 and 7 filter the output data. This erodes away small regions of black pixels and then fills back the region where pixels remain. This eliminates regions with sparse black pixels and thin black lines. The next step eliminates any remaining stray readings. Lines 8-16 examines all the remaining black regions in the image and selects only the largest such region. The resulting output is shown in Figure~\ref{fig:code_flowchart}(c). If this particular region is larger than the \texttt{min\_area} threshold, the function returns the region description (contour, center coordinates and area). 

\begin{figure}[t]
\begin{lstlisting}[language=Python]
# find the colored regions based on HSV range 
def FindColor(imageHSV, lower_col, upper_col, min_area):
    # find the colored regions
    mask=cv2.inRange(imageHSV,lower_col,upper_col)
    # this removes noise by eroding and filling in
    maskOpen=cv2.morphologyEx(mask,cv2.MORPH_OPEN,kernelOpen)
    maskClose=cv2.morphologyEx(maskOpen,cv2.MORPH_CLOSE,kernelClose)
    conts, h = cv2.findContours(maskClose, cv2.RETR_EXTERNAL, cv2.CHAIN_APPROX_NONE)
    # Finding bigest  area and save the contour
    max_area = 0
    for cont in conts:
        area = cv2.contourArea(cont)
        if area > max_area:
            max_area = area
            gmax = max_area
            best_cont = cont
    # identify the middle of the biggest  region
    if conts and max_area > min_area:
      M = cv2.moments(best_cont)
      cx,cy=int(M['m10']/M['m00']),int(M['m01']/M['m00'])
      return best_cont, cx, cy, max_area
    else:
      return 0,-1,-1,-1
 \end{lstlisting}
   \caption{Code to find largest region that matches color.}
  \label{fig:code_colors}
\end{figure}
 
 Once I have the contour of the largest black region (in a variable called \texttt{best\_blackcont}), I use OpenCV to find the best fit rectangle to the area using the following code:

\begin{lstlisting}[language=Python]
blackbox = cv2.minAreaRect(best_blackcont)
\end{lstlisting}

The output of this is shown in Figure~\ref{fig:code_flowchart}(d). The green outline shows that the best fit rectangle closely fits the line in front of the vehicle. This best fit rectangle may be in any rotation on the screen. 

The next step is to compute how far the vehicle is from the line to determine how to steer the vehicle. This is done by looking at the coordinates of the bottom left corner of the best fit rectangle (\texttt{x\_min} below) and determining how far it is from the center of the screen ($x coordinate = 160$). This value ranges from 160 to -160; I multiply this by 0.333 to scale this error estimate to a smaller range that is appropriate for the motor control computations performed in the next step. This is done with the code below:

\begin{lstlisting}[language=Python]
(x_min, y_min), (w_min, h_min), lineang = blackbox
distance = 0.333*(160-x_min)
\end{lstlisting}

This provides an instantaneous error estimate that can be used in a feedback control system. I choose to use a Proportional-Integral-Derivative (PID) controller to provide robust and accurate road-following behavior under a wide range of conditions. Equation~\ref{eq:1} represents the PID implementation, with a decay rate for the integral. While the proportional part of the control system relies only on the instantaneous error, or displacement, I also need to compute the derivative and integral of the error to support PID.  The integral in PID is meant to provide a summary of past behavior of the system (i.e. has it been following the road accurately or drifting continuously to one side or the other). For the derivative, I compute the change in error by keeping track of the previous error reading.

The main problem when implementing integrals and derivatives of sensor data is that they must be approximated as seen in equation~\ref{eq:2}. I use Riemann sums~\cite{calculus} of all the past errors to approximate the definite integral of the vehicle's displacement. The derivative is approximated by finding the change in error. The angular displacement, $\theta$, can be used in place of the derivative when calculable, as seen in equation~\ref{eq:3}.

\begin{equation} \label{eq:1}
C(t_c) = K_p x(t_c) + K_i\int_0^{t} 0.9^{t-t_c}x(t)\mathrm{d}t + K_d\frac{{d}x}{{d}t} 
\end{equation}
\begin{equation} \label{eq:2}
C(n) \approx K_p x_n + K_i(\displaystyle\sum_{i=1}^{n} 0.9^{i-n}x_i ) + K_d(x_n - x_{n-1}) 
\end{equation}
\begin{equation} \label{eq:3}
C(n) \approx K_p x_n + K_i(\displaystyle\sum_{i=1}^{n} 0.9^{i-n}x_i ) + K_d(\theta)
\end{equation}

Once the error, integral and derivative are computed, they are combined using different weighting factors ($K_p, K_d$ and $K_i$, respectively) to produce a correction estimate. This correction value is used to control the relative power of the two wheels - thus, steering the vehicle. Determining these weighting factors is one of the more complex steps in creating an accurate PID control system. I use the Ziegler-Nichols method~\cite{1942.Ziegler} for tuning PID controllers. I begin by setting $K_d$ and $K_i$ to 0 and adjusting $K_p$ until the vehicle followed the road accurately. I then adjust $K_i$ to improve performance, especially along the turns in the road. Finally, I adjust $K_d$ to minimize oscillations. These adjustments are done both based on visual observation and by graphing/analyzing the recorded error values to see that there is a quantifiable decrease in error at each tuning step. 

\begin{figure}[t]
  \centering
  \includegraphics[width=\linewidth]{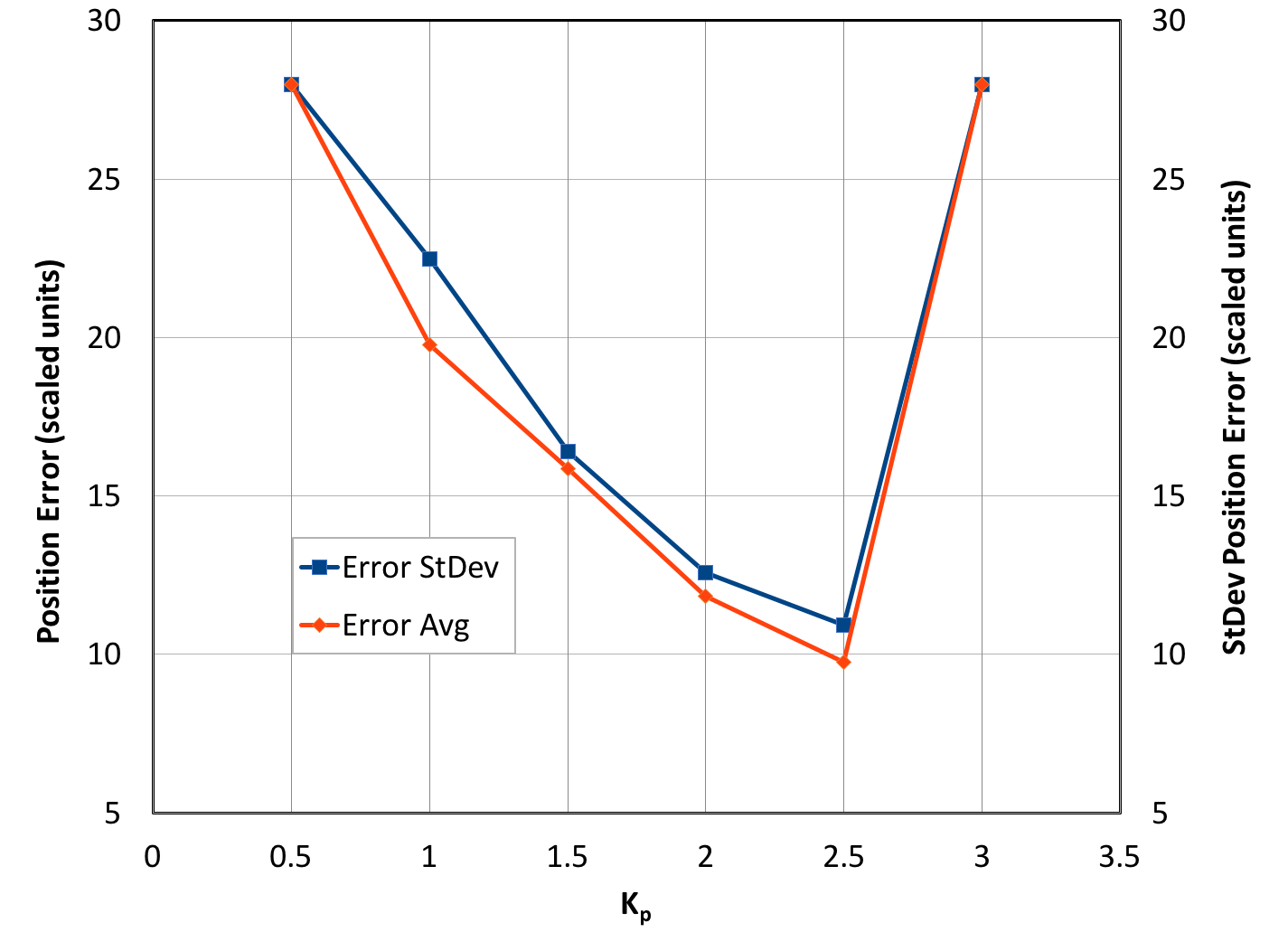}
  \caption{In this experiment, $K_i$ and $K_d$ are set to 0. $K_p$ is varied from $0.5$ to $3.0$ and the average and standard deviation of the position error of the vehicle are plotted above. }
  \label{fig:p_tuning}
\end{figure}

Figure~\ref{fig:p_tuning} shows the first step of this process and how driving quality, plotted on the y-axis using metrics described in Section~\ref{sec:results}, varies as I adjust $K_p$. Based on this graph, I chose $K_p = 1.5$. Note that while this does not produce the optimal, minimum error v
.0alues, the choice of $K_p = 1.5$ provides a combination of low error and stable behavior. In contrast, chosing $K_p = 2.5$ may reduce error but slightly higher values of $K_p$ result in unstable vehicle behavior and high error. 

\begin{figure}[t]
  \centering
  \includegraphics[width=\linewidth]{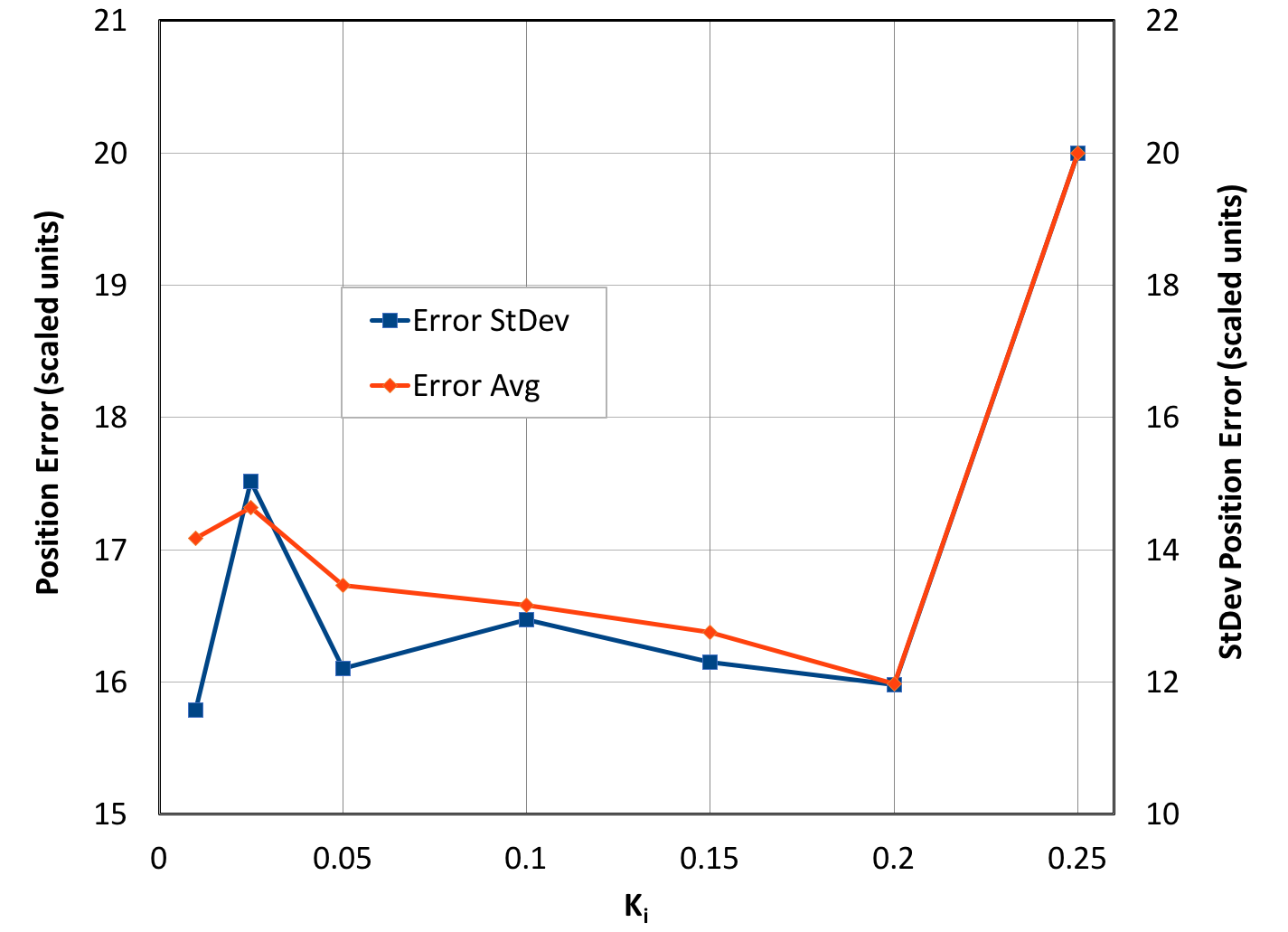}
  \caption{In this experiment, $K_p$ is set to 1.5 and $K_d$ is set to 0. $K_i$ is varied from $0.01$ to $0.25$ and the average and standard deviation of the position error of the vehicle are plotted above. }
  \label{fig:i_tuning}
\end{figure}

In tuning $K_i$ in the second step, I find that at even small values of $K_i$, the driving is unstable as the integral value increases significantly when the vehicle goes around corners.  
I find that this because the simple accumulator suffers from a problem called integrator windup~\cite{PIDwindup}. To address this issue, I add an exponential decay component by multiplying the previous sum by $0.9$ to limit the amount of history that the integral accumulates. Figure~\ref{fig:i_tuning} shows the the impact of $K_i$ on driving quality with $K_p$ set to $1.5$ and the decay component added. Based on this measurement, I chose $K_i$ to be $0.15$ since it produces close to the minimum error values while providing stable performance. As with tuning $K_p$, values that produce slightly lower error (e.g. $K_i = 0.2$) are close to values that produce significantly higher error (e.g. $K_i = 0.25$) and are therefore less stable or reliable. 

\begin{figure}[t]
  \centering
  \includegraphics[width=\linewidth]{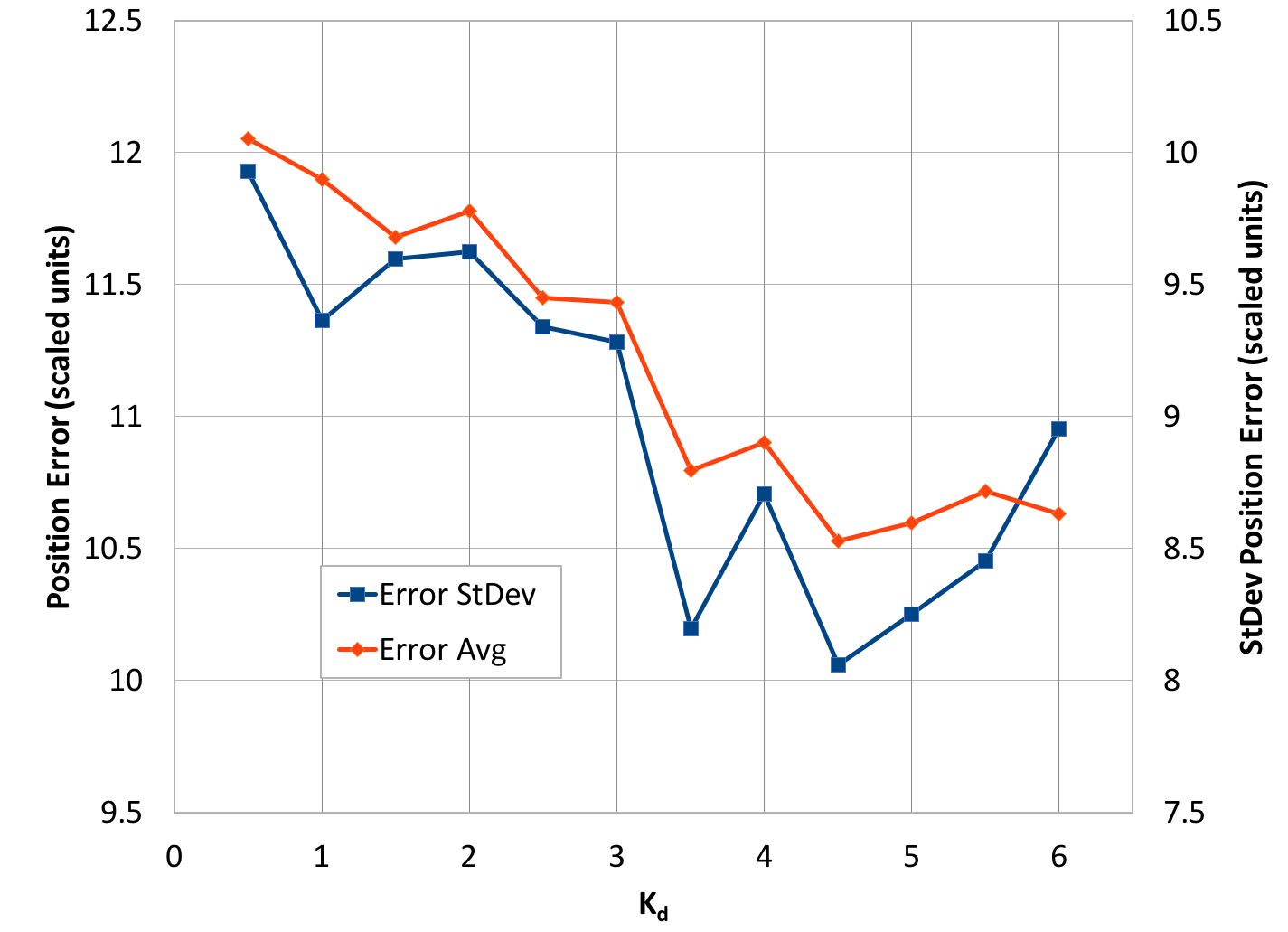}
  \caption{In this experiment, $K_p$ is set to 1.5 and $K_i$ is set to 0.15. $K_d$ is varied from $0.5$ to $6.0$ and the average and standard deviation of the position error of the vehicle are plotted above. }
  \label{fig:d_tuning}
\end{figure}

Finally, Figure~\ref{fig:d_tuning} shows the impact of changing $K_d$ on driving quality, with $K_p$ set to 1.5 and $K_i$ set to 0.15. Based on this measurement, I set $K_d = 4.5$ since it produces both minimal error and stable behavior.  

The end result of tuning the PID controls for the on-vehicle camera is that $K_p = 1.5, K_d = 0.15$ and $K_i = 4.5$ and a decay constant of $0.9$ for the integral produced accurate and robust driving performance. The same tuning methodology is used to tune both infrastructure sensors to produce similar results.
The code for these PID computations is shown below:

\begin{lstlisting}[language=Python]
error = distance
integral = error + 0.9*integral
derivative = error - last_error
last_error = error
Kp = 1.5
Kd = 4.5
Ki = 0.15
correction = Kp*error + Kd*derivative + Ki*integral
left = int(100 - correction)
right = int(100 + correction)
\end{lstlisting}

In using multiple sensors, I realized that not all sensor views were equal. For example, a view from far away provides a less accurate position estimate for the vehicle and road. Similarly, a view from the car that is partially blocked could be similarly inaccurate. As a result, I choose to add a confidence value to sensor analysis feeds from the on-vehicle and infrastructure sensors. In the case of this on-vehicle camera, I find that the area of the rectangle around the road provides useful feedback. When the rectangle is small, it indicates that the code is unable to find the road and the confidence is low. Based on this observation, I compute a confidence value by scaling the size of the black contour area as shown here:

\begin{lstlisting}[language=Python]
confidence = blackarea/55
\end{lstlisting}

The final step of this program transmits the left power, right power and confidence values to the vehicle control software. Based on the results of Phase I of this project, I use the UDP network protocol instead of the TCP network protocol to transmit these values since it results in a much more robust overall system. 
After transmitting, the system immediately continues to the next video frame to provide control feedback as frequently as possible. My measurement of the processing of this sensor feed shows that the Raspberry Pi Model 3B+ is able to process 11 frames per second.

\subsection{Infrastructure Sensor Processing}
\label{sec:infrastructure}

\begin{figure}[t]
  \centering
  \includegraphics[width=\linewidth]{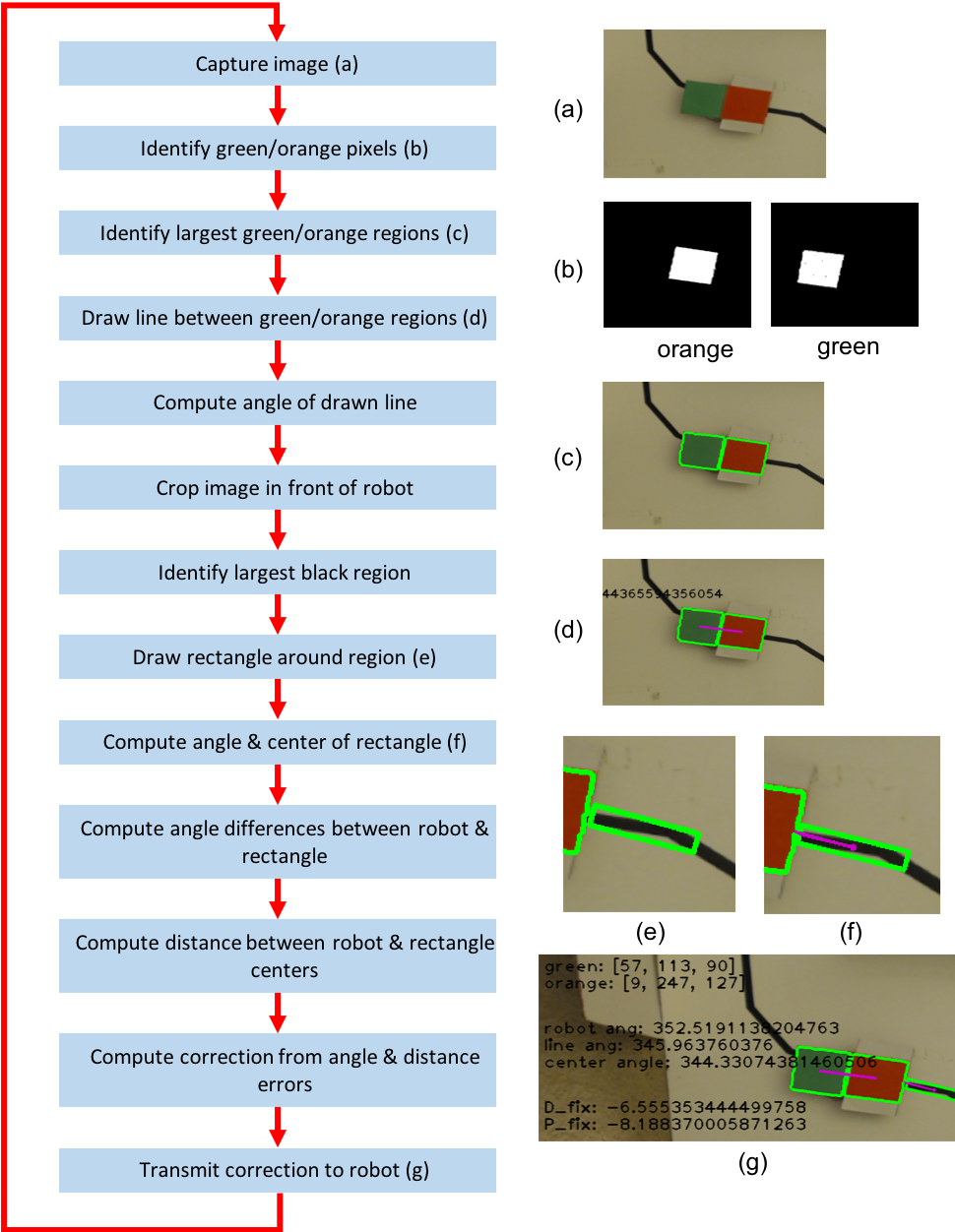}
  \caption{A flowchart of the infrastructure sensor processing code.}
  \Description{The 1907 Franklin Model D roadster.}
  \label{fig:code_flowchart2}
\end{figure}

The objective of the infrastructure sensor processing is similar to that of the on-vehicle processing. It must analyze a video feed to determine the relative location of the vehicle and the road; and use this estimate to compute steering corrections and confidence values that it transmits to the vehicle control software. The flow chart of this processing is shown in Figure~\ref{fig:code_flowchart2} and is slightly more complex than the on-vehicle processing. There are two reasons for this: 1) the infrastructure must determine the vehicle's position whereas the on-vehicle processing implicitly knows this and 2) the infrastructure has a better view of the road ahead and can analyze it in more sophisticated fashion. I describe each of steps in the infrastructure processing in detail below.

As with the on-vehicle processing, the first step is to resize the image from the high quality ($1920\times1080$ resolution) to a lower resolution ($1280\times720$) and convert the image representation to use HSV (hue, saturation and value). The code below performs this capture and conversion and the output is shown is Figure~\ref{fig:code_flowchart2}(a).
 
\begin{lstlisting}[language=Python]
ret, cap_img=cam.read()
img=cv2.resize(cap_img,(1280,720))
imgHSV = cv2.cvtColor(img,cv2.COLOR_BGR2HSV)
\end{lstlisting}

One important difference from the on-vehicle processing is the final resolution used ($1280\times720$ vs. $320\times280$). This is because of the significantly greater computer processing power available for image processing in the infrastructure - a Intel i7 based PC vs. a Raspberry Pi. While future vehicles may have more dedicated compute power for this purpose, it remains likely that the infrastructure will always have more compute resources than the vehicle. 

The next several steps identify the vehicle within the image. Note that there is no equivalent step in the on-vehicle processing. However, the \texttt{FindColor} (Figure~\ref{fig:code_colors}), which is used to find the black road in on-vehicle processing, plays a similar significant role in finding the green region associated with the vehicle. Once the green region is found, the system crops the image to a small area around the found green region and uses \texttt{FindColor} to search for orange within this area. This reduces the likelihood of finding an incorrect orange region. The processing within \texttt{FindColor} is shown in Figure~\ref{fig:code_flowchart2}(b) and its output with the found green and orange regions is shown in Figure~\ref{fig:code_flowchart2}(c). The code to search for these regions is shown below. 

\begin{lstlisting}[language=Python]
# find largest green region
best_greencont, greencx, greency, greenarea = FindColor(imgHSV, lower_green, upper_green, 3000, "green")
# crop frame to be around robot only
robotimgHSV = imgHSV[max(greency-200,0):greency+200,max(greencx-300,0):greencx+300]
# find orange region in this cropped area
best_orangecont, orangecx_incrop, orangecy_incrop, orangearea = FindColor(robotimgHSV, lower_orange, upper_orange, 3000, "orange")
\end{lstlisting}

Once I have the center of the green and orange regions of the robot, I can compute its orientation with the code using the \texttt{ComputeRobotAngle} function shown in Figure~\ref{fig:code_findangle}. Note that this code has to deal with a few special cases. First, it handles the discontinuity of tangent/arctangent at 90 and 270 degrees. Second, arctangent only returns values from 90 to -90 degrees. However, I need the orientation of the robot in 0 to 360 degrees. The code first determines if the robot is facing left or right by comparing the green and orange center X coordinates. It also adjusts negative readings to the 270 to 360 degree range. 

\begin{figure}[t]
\begin{lstlisting}[language=Python]
def ComputeRobotAngle(greencx, greency, orangecx, orangecy):
    # find the angle from the center of green to center
    # of orange this is the angle of the robot in the image
    # I need to special case of 90/-90 due to tan()
    # discontinuity I also need to deal with angles > 90
    # and < 0 to map correctly to a 360 degree circle
    if (greencx-orangecx) == 0:
        if greency > orangecy:
            ang = 90
        else:
            ang = 270
    else:
        ang = 180/np.pi * np.arctan(float(orangecy-greency)/float(orangecx-greencx))
        if greencx > orangecx:
            ang = 180 + ang
        elif ang < 0:
            ang = 360 + ang
        ang = 360-ang
    return ang
\end{lstlisting}
   \caption{Code to find largest region that matches color.}
  \label{fig:code_findangle}
\end{figure}

At this point, I have the robot's location and orientation. In the next few steps, I determine the location and angle of the black line in front of the robot. I begin by locating the middle of the leading edge of the robot. Since the orange and green regions are of the same size, I can use the orange and green box measurements determined above to obtain this location. From this location, I crop a small portion of the overall image. This is done by the Python code below.

\begin{lstlisting}[language=Python]
ylen = (greency-orangecy)
xlen = (greencx-orangecx)
boxX = orangecx - xlen/2
boxY = orangecy - ylen/2
cropHSV=imgHSV[int(abs(boxY-cropsize)):int(abs(boxY+cropsize)),
               int(abs(boxX-cropsize)):int(abs(boxX+cropsize))]
\end{lstlisting}

The coverage of the cropped image is seen in Figure~\ref{fig:code_flowchart2}(e). Within this portion of the image, I look for the black pixels and identify the largest black region using \texttt{FindColor}. Once I have the contour of the largest black region (in a variable called \texttt{best\_blackcont}), I use OpenCV to find the best fit rectangle to the area using the following code:

\begin{lstlisting}[language=Python]
blackbox = cv2.minAreaRect(best_blackcont)
\end{lstlisting}

The output of this is shown in Figure~\ref{fig:code_flowchart2}(e). The green outline shows that the best fit rectangle closely fits the line in front of the vehicle. This best fit rectangle may be in any rotation on the screen. 

Next, the program uses OpenCV to obtain the angle that the rotation angle of the best fit rectangle. Unfortunately, the range of angles is only 0 to -90 degrees, indicating the rotation of the side closest to horizontal. In the code below, I use the fact that the line is thin (i.e. it is longer than it is wide) and that the vehicle is facing a particular direction (the \texttt{ang} variable below) to map this 0 to -90 reading to the correct quadrant. This produces a 0 to 360 degree reading (in the variable \texttt{lineang}) that can be compared with the direction that the vehicle is moving. The below code also identifies the center of the line in the cropped region of the image. The estimate of the line's center and angle are shown in the violet line in Figure~\ref{fig:code_flowchart2}(f).

\begin{lstlisting}[language=Python]
(x_center, y_center), (width, height), lineang = blackbox
if width > height:
    if (ang > 135):
        lineang = 180 - lineang
    else:
        lineang = -1 * lineang
else:
    if (ang > 270) or (ang < 45):
        lineang = 270 - lineang
    else:
        lineang = 90 - lineang
\end{lstlisting}

I now have the direction and center black line in front of the vehicle, as well as the direction and location of the center of the front. I use these measurements to compute two values that summarize the vehicle's current navigation status. First, I compute the difference between the vehicle's movement direction and the direction of the line and store it in the variable \texttt{D\_fix} (in short for ``direction fix'') below. 

\begin{lstlisting}[language=Python]
D_fix = lineang - ang
\end{lstlisting}

Ideally this should be 0 if the vehicle is navigating accurately on the line. Note that this value would be 0 even if the vehicle was navigating parallel to the line. For this reason, I need a second value to represent how far the vehicle is offset from the line. To do this, I draw a line from the center of the vehicle's front (i.e. the center of the orange region's leading edge in Figure~\ref{fig:code_flowchart2}(e)) to the center of the line (i.e. the start of the violet line in Figure~\ref{fig:code_flowchart2}(f)). In the below (\texttt{x\_center, y\_center}) are the coordinates of the line's center and (\texttt{r\_center, r\_center}) are the coordinates of the vehicle's center. As with the earlier computations, in converting (x, y) coordinate measurements to angles, I have to handle discontinuities in the arctangent function and the fact that it only provides output values of 90 to -90 degrees. The measurement of this line's angle is stored in the variable \texttt{offset\_ang}. The angle of this line relative to the vehicle's motion represents the left to right offset of the line from the vehicle's center. I store this difference in the variable \texttt{P\_fix} (short for ``position fix'') \\
\\
\begin{lstlisting}[language=Python]
if (x_center - r_center) == 0:
    if (ang < 180):
        P_fix = 90 - ang
    else:
        P_fix = 270 - ang
else:
    offset_ang=180/np.pi * np.arctan((r_center-y_center)/(x_center-r_center))
    if (offset_ang < 0):
        if (ang > 225):
            offset_ang = 360 + offset_ang
        else:
            offset_ang = 180 + offset_ang
    elif (ang > 135 and ang < 315):
            offset_ang = 180 + offset_ang
    P_fix = offset_ang - ang
\end{lstlisting}

After all the image processing is complete, I have two variables, \texttt{P\_fix} and \texttt{D\_fix}, that represent the vehicle's offset from the line and relative angle to the line. These are equivalent to the error and derivative estimates that are commonly used in the PID (Proportional-Integral-Derivative) algorithms used in numerous control systems. \texttt{P\_fix} represents the current error -- i.e. how far the vehicle is from its target position and \texttt{D\_fix} represents the rate at which the error is changing -- i.e. the slope between the movement and the black line. Note that the infrastructure sensor code can compute this derivative of the error from the image itselft since it has a more clear view of the road's direction than the on-vehicle sensors. To support the third component of a PID control system, the program computes an integral using the same technique as the on-vehicle code -- by adding the error to the past integral and using a $0.9$ factor exponential decay to prevent integral windup:


The code for the PID computation (shown below) is similar to the on-vehicle code. I use the Ziegler-Nichols method~\cite{1942.Ziegler} for tuning this PID controllers as well. Note that the resulting $K_p, K_d$ and $K_i$ values differ from those used by the on-vehicle code. This is for a variety of reasons including differences in communication delays to the vehicle control code and difference in measurements of both error and derivative.  I have omitted the graphs associated with tuning this PID for readability since it is similar to the tuning in Section~\ref{sec:on-vehicle}.

\begin{lstlisting}[language=Python]
Kp = 1.0
Kd = 0.5
Ki = .02
correction = Kp*P_fix + Kd*D_fix + Ki*integral
left = int(100 - correction)
right = int(100 + correction)
\end{lstlisting}

While these PID weights differ between the infrastructure and on-vehicle code, the resulting output (i.e. the left and right motor powers) are conceptually similar and can be combined if needed. As mentioned earlier, I choose to add a confidence value to sensor analysis feeds from the on-vehicle and infrastructure sensors. In this case, the size of the black region identified as the road in front of the car provides a reasonable confidence estimate. When the rectangle is small, it is typically because the system was unable to find the road segment and is an inaccurate estimate. Based on this observation, I compute a confidence value by scaling the size of the black contour area by the optimal possible size of the area (13), as shown below. While this uses a similar technique as the on-vehicle confidence estimate, I find that the scaling factor needs to be different.

\begin{lstlisting}[language=Python]
confidence = blackarea/13
\end{lstlisting}

In the final step, the left motor power, right motor power and confidence are sent to the vehicle control software using a UDP socket.

\subsection{Vehicle Control Software}
\label{sec:vcs}

The vehicle control software is responsible for collecting the steering commands from all sources (on-vehicle and multiple infrastructure cameras) and combining them to determine the appropriate power to give each motor. Combining sensor values, also known as sensor fusion~\cite{fusion}, can be performed using a number of algorithms. There are two components of my sensor processing design that significantly help in this sensor fusion design: a common scale and confidence values. 

\paragraph{Common Scale} The output of the sensor processing systems described in Sections~\ref{sec:on-vehicle}~and~\ref{sec:infrastructure} comes in the form of motor power control. Therefore, even if each system measures different properties or has different views, the software associated with the sensor feed must convert this to a common scale. For example, if the on-vehicle software uses a combination of GPS localization and radar sensors, while the infrastructure systems use cameras, they both output left and right motor power as the final decision. This makes it possible for the vehicle control software to combine these outputs using simple mathematical operations such as average or maximum instead of more complex sensor-specific algorithms. 

\paragraph{Confidence Values} As described in Sections~\ref{sec:on-vehicle}~and~\ref{sec:infrastructure}, each sensor feedback report sent to the vehicle control software comes with an associated confidence values. This provides some indication on the accuracy of the sensor readings that produce the transmitted result. This enables the vehicle control software to weight the different inputs when combining them.

One of the primary goals of this project was to understand how to combine the sensor readings from different sources. To achieve this goal, I tested three different sensor fusion algorithms:
\begin{enumerate}
    \item {\bf Maximum Confidence.} Here, the sensor feedback with the highest confidence is used to steer the vehicle. All other feedback is simply discarded.
    
    \item {\bf Simple Average.} The sensor feedback is combined using a simple numeric average of the steering feedback. The confidence information is not used and the average is calculated as follows:
    
        \begin{equation} \label{eq:4}
        simple~average = \frac{\displaystyle\sum_{i=1}^{n}r_i}{n}
        \end{equation}

    where $r_i$ represents the reading from the sensor $i$ and there are $n$ sensors in total.

    \item {\bf Confidence-weighted Average.} This average is computed with each sensor's feedback weighted by its associated confidence. Mathematically, it is computed as follows
        \begin{equation} \label{eq:5}
        weighted~average = \frac{\displaystyle\sum_{i=1}^{n}c_i \times r_i}{\displaystyle\sum_{i=1}^{n}c_i}
        \end{equation}

     where $r_i$ represents the reading from the sensor $i$, $c_i$ represents its confidence and there are $n$ sensors in total.
This is implemented by the following code:
\begin{lstlisting}[language=Python]
lmotor_speed = (picam_left * picam_confidence +
        + cam0_left * cam0_confidence 
        + cam1_left * cam1_confidence) /
        (picam_confidence + cam0_confidence + cam1_confidence)
        
rmotor_speed = (picam_right * picam_confidence +
        + cam0_right * cam0_confidence 
        + cam1_right * cam1_confidence) /
        (picam_confidence + cam0_confidence + cam1_confidence)

LMotor.setSpeed(lmotor_speed)
RMotor.setSpeed(rmotor_speed)
\end{lstlisting}
\end{enumerate}

Note that this is not an exhaustive set of sensor fusion algorithms. For example, I could have used Kalman Filters~\cite{kalman} or Particle Filters~\cite{particle} where the vehicle's position would be updated in each step by the velocity and trajectory estimates and then filtered by the respective sensor observations. While these approaches could provide greater accuracy in the vehicle's position estimate, the above techniques provide more direct insight into how the different treatment of confidence impact driving performance. This insight would be needed to make an effective Kalman or Particle filter as well. 
\section{Experimental Results}
\label{sec:results}

The goal of this project was to explore the benefits of using both on-vehicle and infrastructure sensing in guiding an autonomous vehicle. 

\paragraph{Metrics} To quantify the benefits of multiple sensors, I first need a way to measure autonomous driving quality. I use two different metrics to measure this:
\begin{itemize}
    \item {\bf Position Error.} My first metric quantifies how close the vehicle is to the center of the road. This distance is measured as part of the image processing code and reported as the current ``error''. However, using the image processing position error value as a metric has some issues. First, when there are multiple cameras in use, they may have different error values that they measure - which one would the correct current distance estimate? Second, even if I choose a particular camera, part of the premise of this project was that there are times when a particular camera does not see the road - what would be the correct value to use at that time? In addition, position error is measured based on camera view; as a result, the units of measurement are relative to the image size and I report this as ``scaled units''
    \item {\bf Correction.} The correction that is computed in the vehicle control software and applied to the motors (i.e. the difference in their power) provides a scaled estimate of how far the system thinks the vehicle is from its correct location. A car that follows the same road with smaller correction values is certainly driving ``better'' than one with larger correction values. One drawback of this metric relative to position error is that a 0 value does not necessarily represent an optimal value. For example, a vehicle that travels along a curved road must apply different power to its wheels (i.e. have a significant correction value) to follow the road and maintain low position error. This may be the optimal value for the vehicle following that path. 
\end{itemize}
For both metrics, higher values indicate worse driving performance. Since each metric has some drawbacks, I use each where most appropriate. To summarize the performance of the vehicle as it travels for some duration, I use the average of the absolute value of these metrics. I use absolute value since these metric values are normally symmetric around the 0 and, therefore, the simple average will be 0. In addition, I record the standard deviation of these readings to quantify how stable the vehicle's movement is. 

Note that anywhere that I report or graph the mean, between 1000 and 6000 samples were used in computing the mean. As a result, the standard error of the mean, which is the standard deviation ($\sigma$) divided by the square root of the number of samples (i.e., $\frac{\sigma}{\sqrt{n}}$), is at most $\frac{1}{30}$th of the standard deviation. Since the standard deviation and average are roughly the same order of magnitude in all my measurements, this indicates that there is little error in the mean and that the measurements of the mean have strong statistical significance. I chose not to include the standard error as an error bar in my graphs to improve readability. 

Experimental evaluation of baseline performance, the impact of sensing failures and the benefit of multiple sensors (on-vehicle and infrastructure) are provided in the following subsections.

\subsection{Baseline Performance}

\begin{figure}[t]
  \centering
  \includegraphics[width=\linewidth]{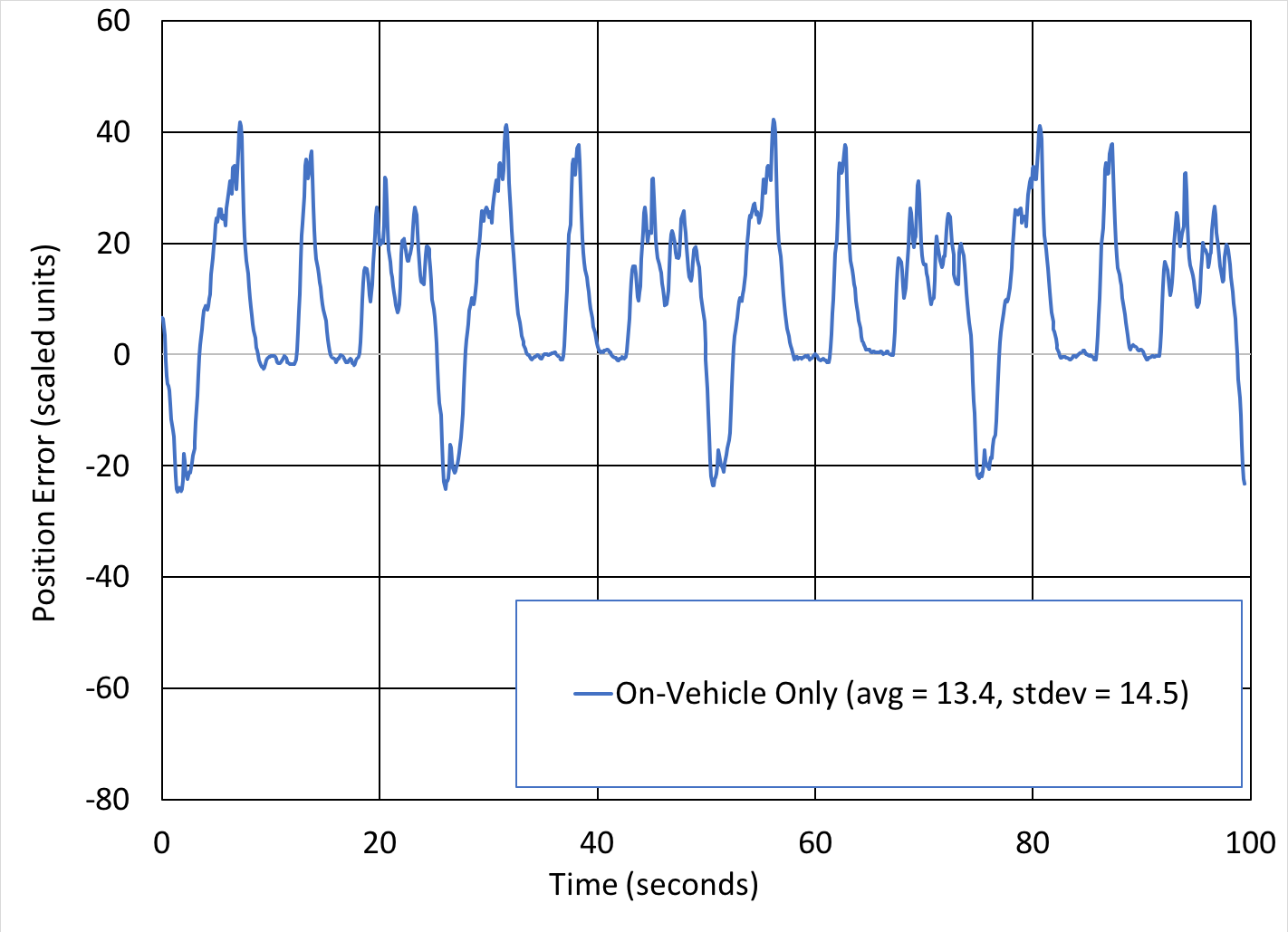}
  \includegraphics[width=\linewidth]{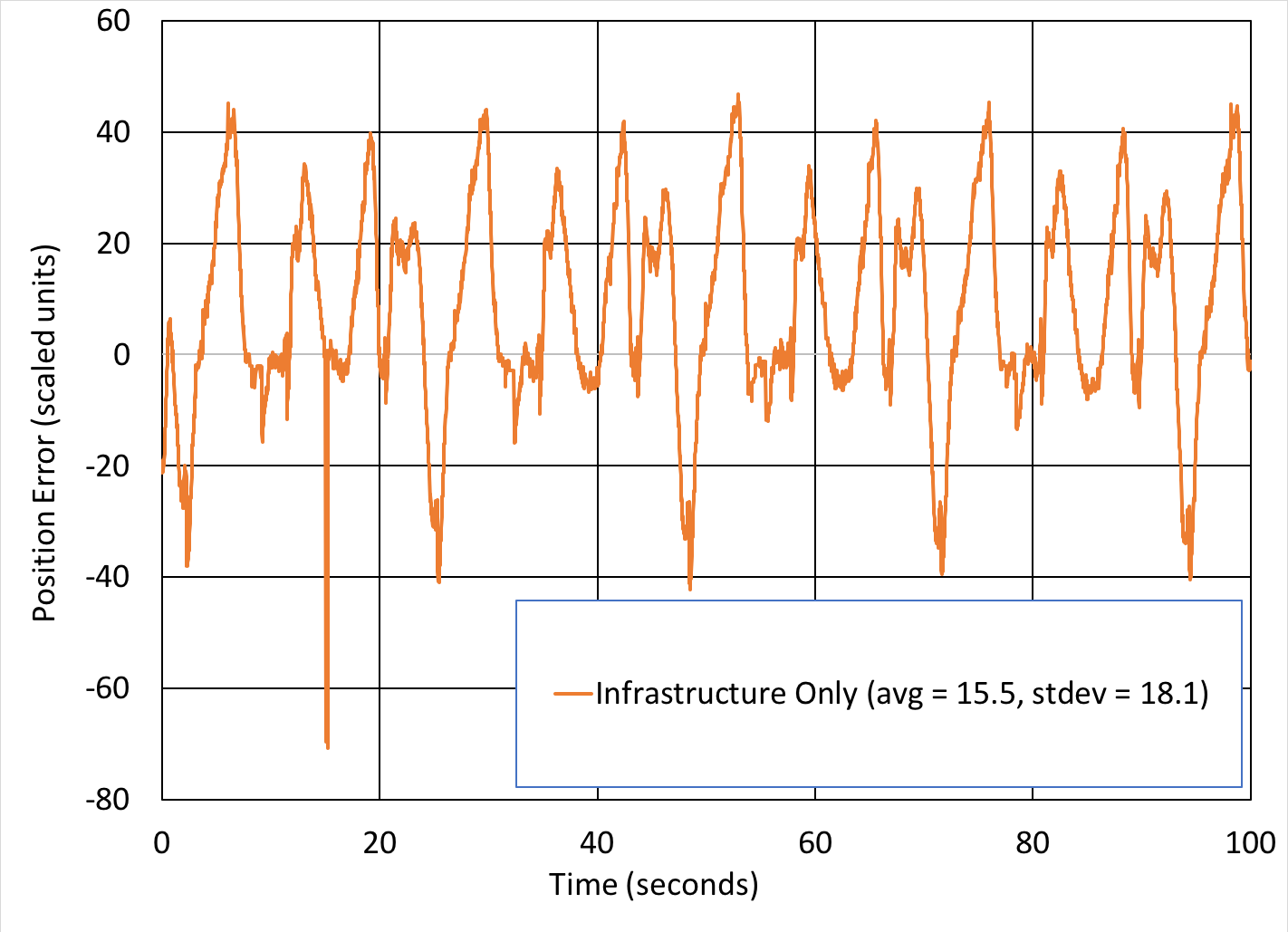}
  \caption{Position error when the system uses only on-vehicle (top) or only infrastructure sensors (bottom). Position error for both approaches are similar with values averaging 13.4 for on-vehicle sensors only versus 15.5 for only using infrastructure sensors.}
  \label{fig:baselinePos}
\end{figure}
\begin{figure}[t]
  \centering
  \includegraphics[width=\linewidth]{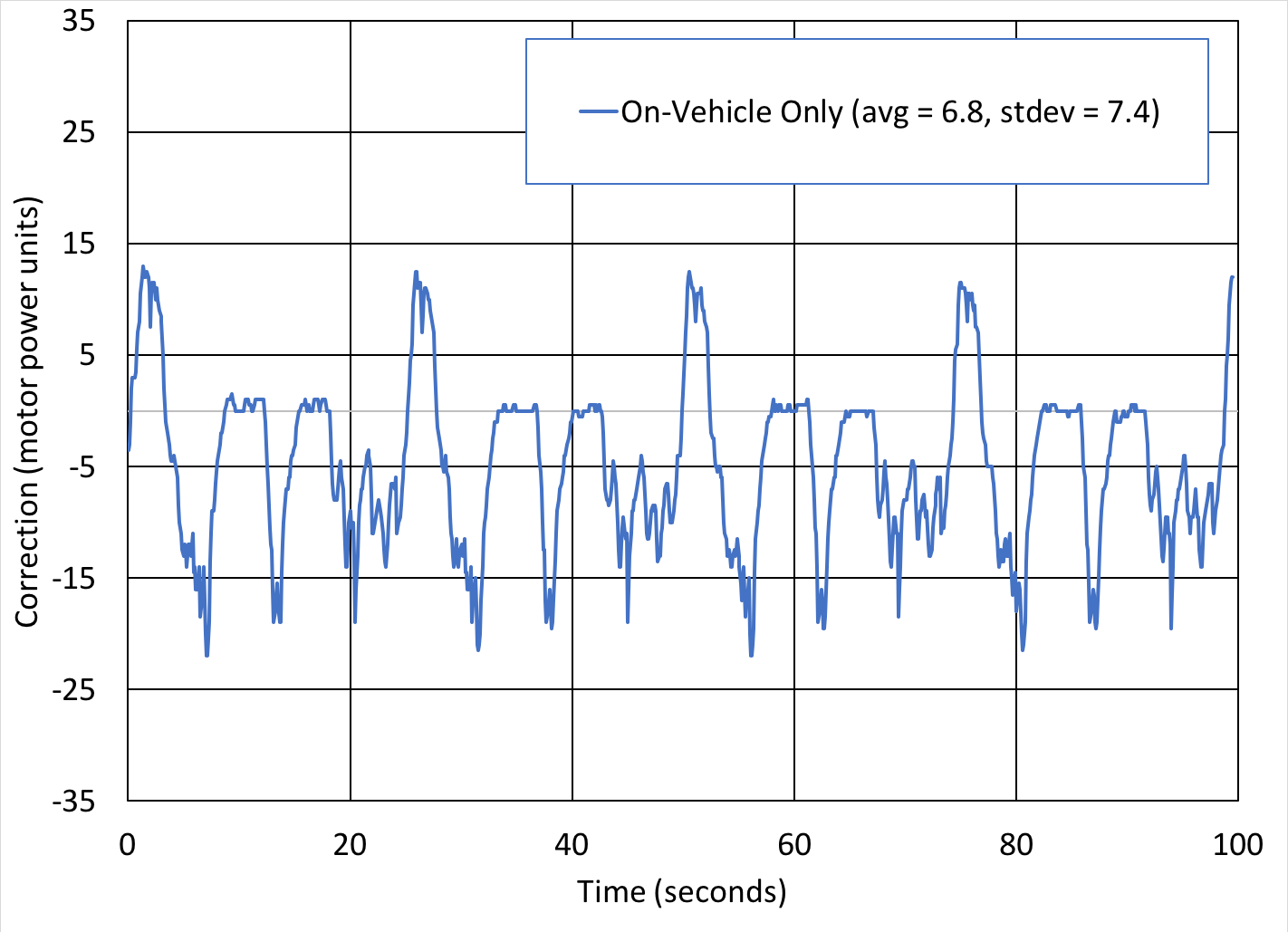}
  \includegraphics[width=\linewidth]{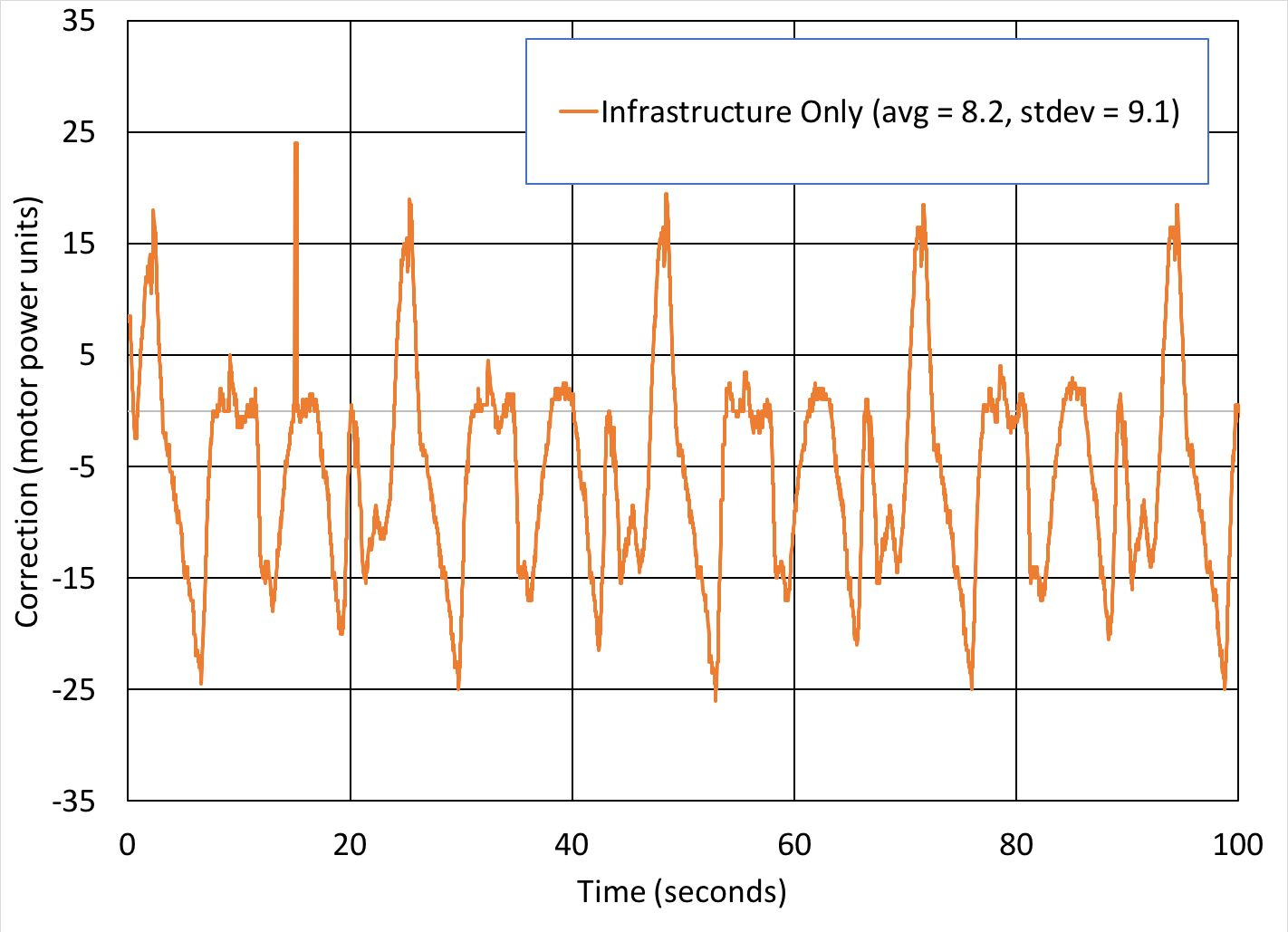}
  \caption{Motor correction when the system uses only on-vehicle (top) or only infrastructure sensors(bottom).Correction values were on average 6.8 when using on-vehicle sensors and 8.2 using infrastructure sensors.}
  \label{fig:baselineCor}

\end{figure}

To understand the benefits of multiple sensor views, I need to first measure the baseline performance of the system with a single sensor view. I ran the vehicle using only the on-vehicle sensor processing or the infrastructure sensor processing. Figure~\ref{fig:baselinePos} shows the reported position error and Figure~\ref{fig:baselineCor} shows the correction as the vehicle travels along the track for 100 seconds. As can be seen from the graphs, the performance of the two approaches is similar. 

Beyond being visually similar, the average position errors are 13.4 and 15.5 for on-vehicle and infrastructure sensing, respectively. Similarly, the standard deviations are 14.5 and 18.1. For the correction values, the averages are 6.8 and 8.2 for on-vehicle and infrastructure sensing, respectively and the standard deviations are 7.4 and 9.1. From these values, it is clear that the two approaches provide similar performance when operating on their own. As a result, the quality of feedback from either system is not clearly superior and, and therefore, does not require any special handling when they are combined through sensor fusion techniques.

\subsection{Impact of Sensing Failures}

\begin{figure}[t]
  \centering
  \includegraphics[width=\linewidth]{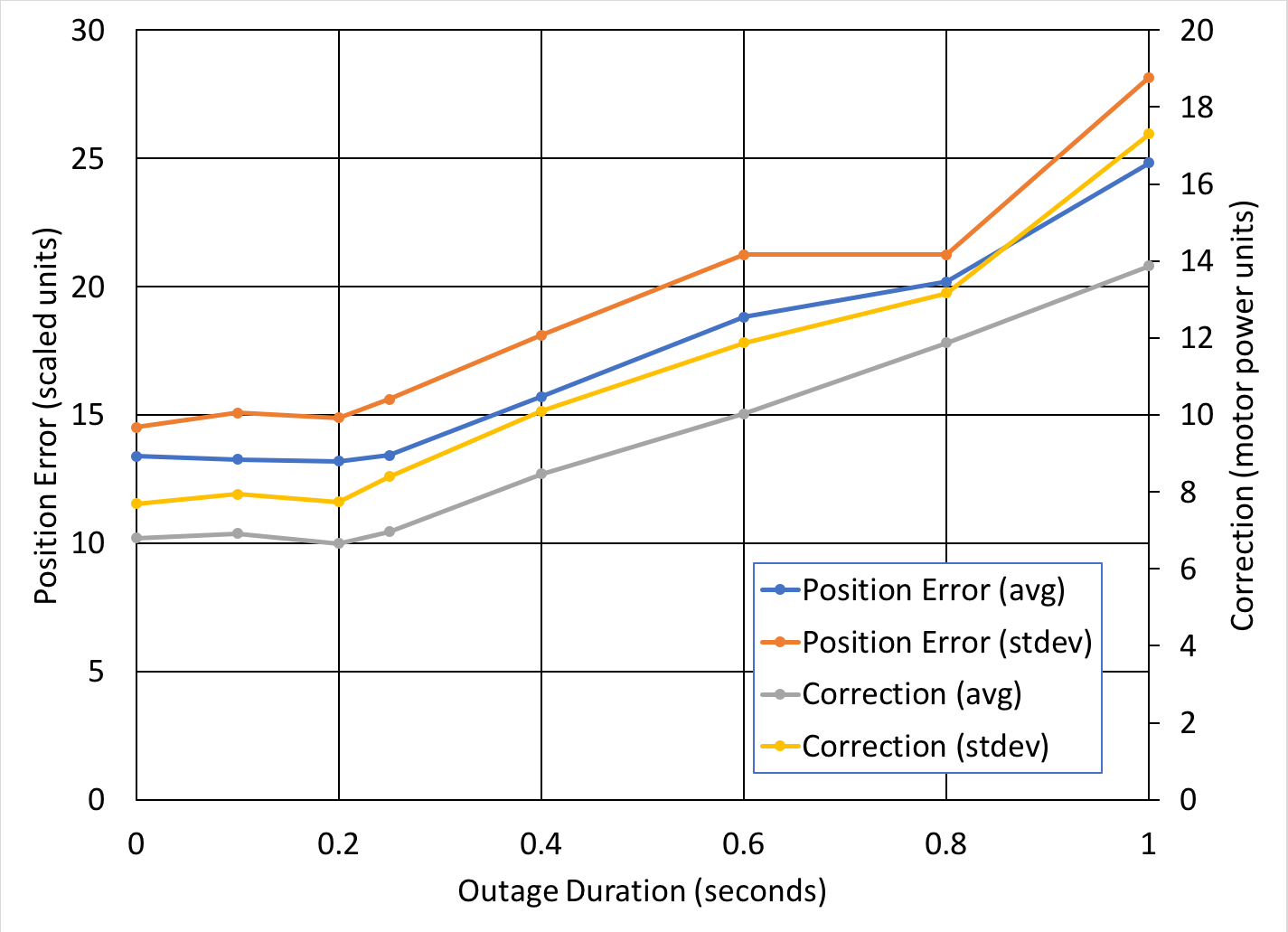}
  \caption{Measurement of driving quality (position error and correction) as outage duration increases. Position Error (both average and standard deviation) are graphed on the left Y axis while Correction (average and standard deviation) are graphed on the right Y axis. }
  \label{fig:outageimact}

\end{figure}

As mentioned in Section~\ref{sec:intro}, current autonomous vehicles suffer from periods during which their sensors are unable to give a reading or give inaccurate readings. To quantify the impact of such sensor outages, I periodically disable a particular sensor and observe the impact it has on driving quality. In this experiment, I have the vehicle navigate using only the on-vehicle sensor. Every 3 seconds, the sensor is disabled for $n$ seconds, where $n$ is varied from 0 to 1 seconds. After each ``outage'', I record the next 5 readings of position error and correction to observe the immediate impact of a sensor feed error on driving quality. In each run of up to 100 seconds, I report the average and standard deviation of these measurement samples. These values are plotted in Figure~\ref{fig:outageimact}. As $n$ increases, you can see a clear decrease in driving quality as both position error and correction increase. This is because the vehicle drives without up-to-date guidance during an outage. Longer durations result in the vehicle moving further from the center of the road. When the sensor feed recovers, the car may be far from the road and take more time to steer back to the proper location. Note that the standard deviations of these metrics also increases with longer sensor outages -- an indicator that driving behavior is more erratic. This is because the vehicle must make abrupt corrections as it tries to recover from the sensor coverage gap. In addition, when the outage was set to 0.8 and 1.0 seconds, the vehicle was unable to stay on the road for the entire 100 second duration and crashed at 30 and 53 seconds into the measurement, respectively. This shows that while the steering control systems is robust, lack of sensor data will make it fail to drive safely. 

\subsection{Benefit of Multiple Sensors}

While I used a periodic approach to introducing sensor failures to quantify the impact of a single sensor outage, I use a probabilistic failure model when there are multiple sensors used in an experiment. This is a more appropriate failure model for evaluation and avoids the necessity for explicitly coordinating the failures of different sensors. In this experiment, each sensor divides time into intervals of 0.4 seconds. The sensors are not synchronized and the intervals on different sensors may start at different times. In each interval, a sensor uses a random number generator to choose a number between 1 and 100. If the chosen number is above a configured probability value $p$, the node transmits actively during the interval. If the number is below this value, it disables the sensor and reports an error value to the 
vehicle control software during the interval. I vary the probability $p$ to examine how robust the system is to sensor failures. 

\begin{figure}[t]
  \centering
  \includegraphics[width=\linewidth]{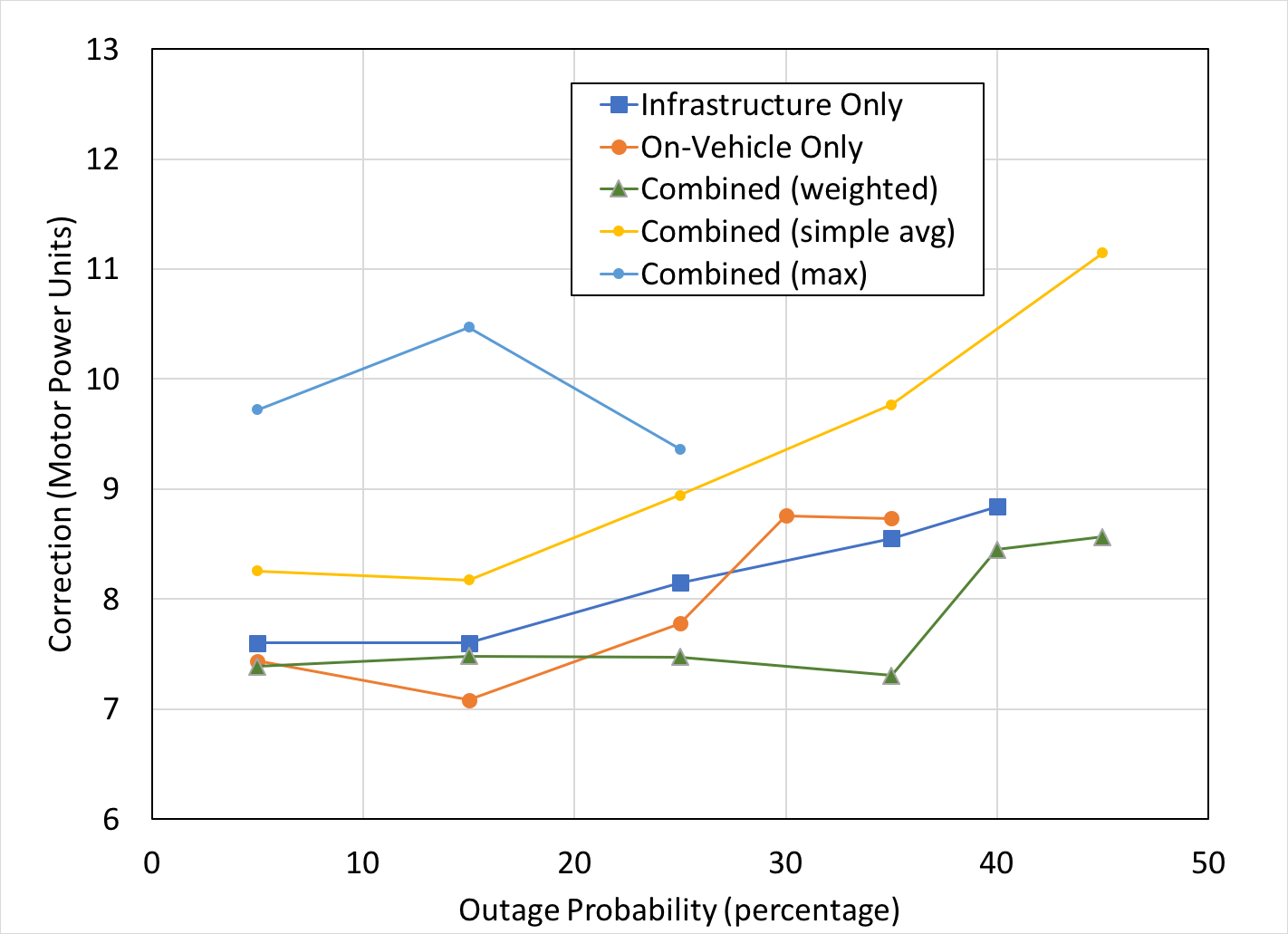}
  \caption{Measurement of driving quality (correction) as outage probability increases. The graph illustrates three different configurations: one that uses just the on-vehicle sensors, one that uses a pair of infrastructure sensors with partially overlapping views and a combined one that uses the on-vehicle and infrastructure sensors. The combined configuration is further divided by the use of three different sensor fusion algorithms: max, simple avg and weighted.}
  \label{fig:benefit}

\end{figure}

\begin{figure}[t]
  \centering
  \includegraphics[width=\linewidth]{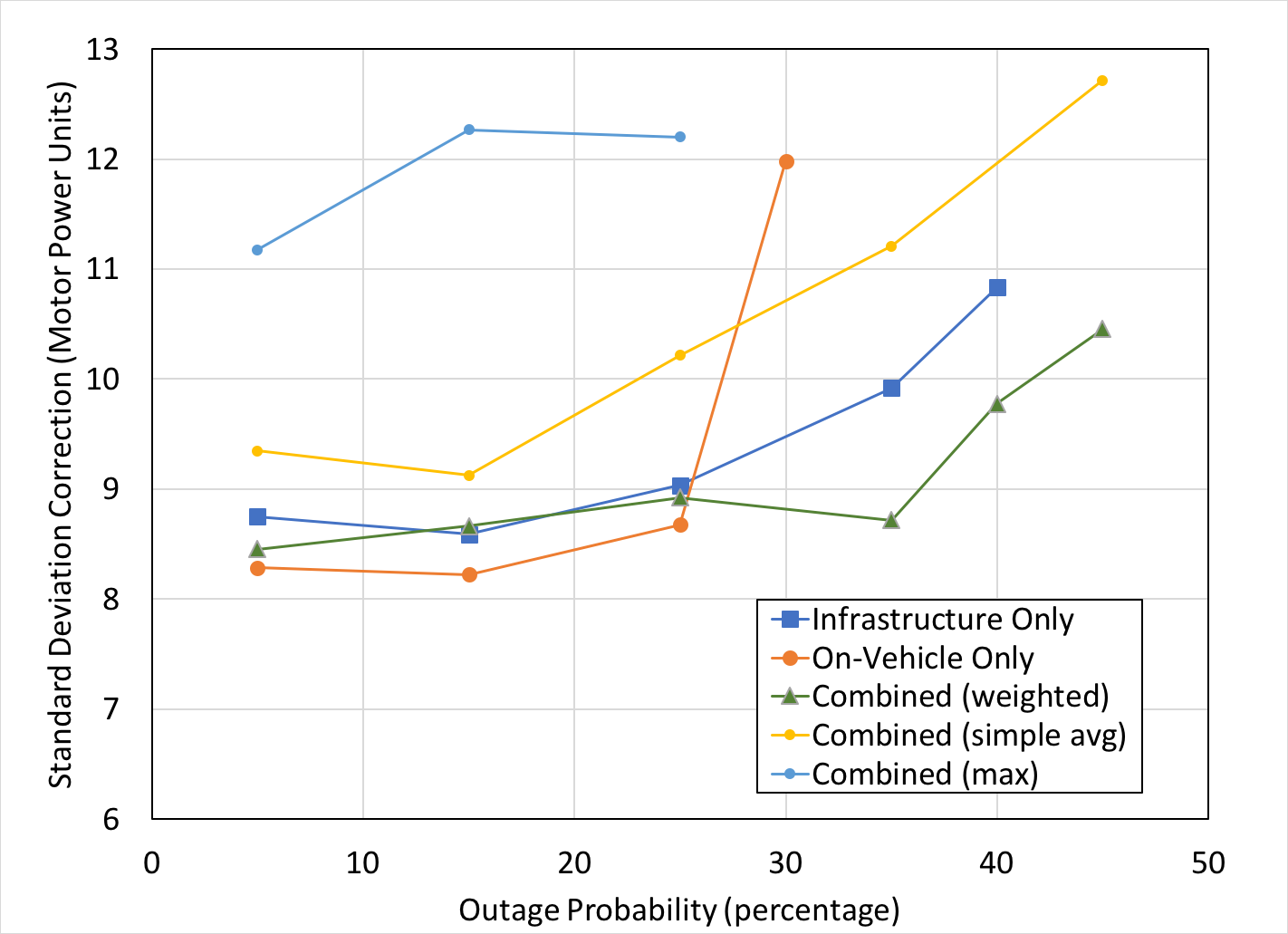}
  \caption{Measurement of driving quality (correction standard deviation) as outage probability increases. Similar to Figure~\ref{fig:benefit}, the graph illustrates three different configurations. The combined approach offers the best solution.}
  \label{fig:benefit2}

\end{figure}

Figure~\ref{fig:benefit} and Figure~\ref{fig:benefit2} plots driving quality using the correction metric as $p$ is varied. The plot shows three different configurations: one that uses just the on-vehicle sensors, one that uses a pair of infrastructure sensors with partially overlapping views and a combined one that uses the on-vehicle and infrastructure sensors. In the combined configuration, three different sensor fusion algorithms are used: Maximum Confidence (max), Simple Average (simple avg) and Confidence-weighted Average (weighted). 

As the graph shows, the on-vehicle configuration is the least robust and the driving quality degrades quickly as the outage probability increases. Once the probability reaches 30\% the vehicle is not able to navigate my road. The pair of infrastructure sensors fares better and is able to navigate up to a 35\% outage probability. The combined approach proves to be highly sensitive to the choice of sensor fusion algorithm. When using the Maximum Confidence algorithm, the vehicle oscillates over the road due to the switching between different sensor feeds. This results in poor driving performance even a low failure rates and low tolerance for failures. When using the Simple Average, the vehicle drives inaccurately due to the feedback from sensors with low confidence observations impacting driving decisions in a significant manner. With the use of Confidence-weighted Averages, the combined configuration provides the best performance and navigates successfully up to a 40\% outage probability. In addition, it is interesting to note that the increase in outage probability has less of an impact on the correction metric when Confidence-weighted Average is used -- for example, the correction metric does not increase even with an outage probability of 35\%. This is because there is frequent feedback from multiple sensors combined with an effective fusion algorithm guiding the vehicle -- providing reliability through redundancy.

\section{Discussion}
\label{sec:discussion}

The experimental study in the previous section enables me to make the follow important observations:

\begin{itemize}
    \item Either on-vehicle and infrastructure sensing can steer an autonomous vehicle with reasonably similar driving quality. 
    \item Even relatively short on-vehicle outages can significantly impact the driving of an autonomous vehicle.
    \item Infrastructure and on-vehicle sensors can be used effectively in combination to create a system that is resilient to sensor reading failures and other issues. 
    \item The choice of sensor fusion algorithm is critical. Both combining all available readings and using confidence values for these readings in the fusion algorithm are necessary to get accurate and robust driving. 
\end{itemize}


The results of my experiments, in both Phase I and Phase II of this project, show that the use of infrastructure sensors has great promise for making autonomous vehicles much safer. Given this observation, I describe below the potential road-blocks that must still be addressed for making this approach practical.

\paragraph{Privacy} The use of cameras throughout our environment to track vehicles will raise numerous privacy concerns. However, it is worth noting that similar camera-based infrastructures are being deployed for other uses. For example, the city of London has 197 cameras around the downtown area that perform automatic number plate recognition to charge automobile owners a toll for vehicle use in the area. New York police have linked 3,000 surveillance cameras with license plate readers, radiation sensors, criminal databases and terror suspect lists. While the existence of current infrastructure does not justify the creation of another system that invades our privacy, these existing deployments do point out that systems that use cameras can be deployed and the data they collect can be managed to avoid major privacy issues. For autonomous driving, the cameras do not need to collect personally identifiable information (PII) such as license plates numbers or faces. In addition, communication between the infrastructure and the vehicles can leverage local wireless links to avoid exposing any information on the wide-area network. 

\paragraph{Coverage} While my design shows the value of infrastructure sensing, the safety gains are only available when the vehicle is within the observation area of infrastructure cameras. As pointed out above, metropolitan areas such as London and New York already have dense coverage. Autonomous vehicles are most likely to be used first in these urban areas. In addition, most challenges facing autonomous vehicles are likely to emerge only in these densely populated urban areas. Driving in rural areas is far more predictable and unlikely to have issues such as pedestrians, bicyclists, frequent stops, cross-walls, small vendor carts, etc. 

\paragraph{Experiment Realism} The simulation and prototype vehicle used in the study may seem unrealistic. The car is identified with a large green and orange identifier. In the real world, automobiles would not be color-coded. However, there are lots of real-world algorithms available today that can identify individual cars without a color code. 

Steering in the autonomous vehicle uses skid-steering. However, how the prototype steers is not a limitation to the actual purpose of the study.

A 2m X 2m field is not realistic. However, the analysis in this paper is about being able to fuse data from various sensors together in real-time to navigate. Once it has been proven to work, I believe that it can be scaled to real-world settings.  

\paragraph{Deployment Incentives} Today, autonomous cars are still in the research stage. Manufacturers such as Uber and Tesla are using only on-vehicle sensors to navigate. This is probably because they want to sell a stand-alone product and do not want to coordinate with government agencies for additional data. Once the market gets more mature and autonomous cars become more common, the safety concerns will only increase. This is where sensor fusion and using infrastructure sensors can play a role.
\section{Conclusion}

My measurements of the impact of sensor failures showed that even short outages (\textasciitilde1 second) at slow speeds (\textasciitilde25 km/hr scaled velocity) prevents vehicles that rely on on-vehicle sensors from navigating properly. However, my results also show that harnessing the thousands of sensor inputs already in place around us could be the answer to how to improve the safety of self-driving vehicles. My system, Horus, shows that it is possible to use sensor fusion to combine on-vehicle sensor data with infrastructure data to significantly improve an autonomous vehicle's ability to reliably navigate. In a scenario with one on-vehicle camera and two infrastructure cameras with overlapping views, Horus could successfully navigate despite sensors being unavailable 40\% of the time. However, to achieve these benefits, the sensor fusion algorithm must be chosen carefully.


\bibliographystyle{ACM-Reference-Format}
\bibliography{references}

\newpage
\onecolumn
\appendix

\section{Code}
\label{sec:code}

\subsection{On-Vehicle Sensor Processing}
\begin{lstlisting}[language=Python]
# import the necessary packages
from picamera.array import PiRGBArray
from picamera import PiCamera
import time,sys
import cv2

import numpy as np
import struct
import socket
import random

# set up network socket/addresses
host = 'localhost'
Lport = 4000
Rport = 5000
sock = socket.socket(socket.AF_INET, socket.SOCK_DGRAM)
sock.setsockopt(socket.SOL_SOCKET, socket.SO_REUSEADDR, 1)
sock.bind(("", Lport))
print ("Active on port: " + str(Lport))
robot_address = (host, Rport)

# initialize the camera
camera = PiCamera()
camera.resolution = (320, 240)
camera.framerate = 32
rawCapture = PiRGBArray(camera, size=(320, 240))
kernelOpen=np.ones((5,5))
kernelClose=np.ones((20,20))
# allow the camera to warmup
time.sleep(0.1)
xdim = 320
ydim = 240
cropsize = 80
gmax=0
rmax=0
lastP_fix = 0
I_fix=0
lower_black=np.array([0,0,0])
upper_black=np.array([180,125,80])

interval = sys.argv[1]
update = sys.argv[1]
#interval = random.randint(1, 10)
duration = sys.argv[2]
threshold = int(sys.argv[3])

def SendToRobot(left, right, error, P, I, D):
    global sock
    data = str(left)+";"+str(right)+";"+str(error)+";"+str(P)+";"+str(I)+";"+str(D)
    send_msg = str(str(data)).encode()
    try:
          sock.sendto(send_msg, robot_address)
          #print send_msg
    except Exception as e:
          print("FAIL - RECONNECT.." + str(e.args))
          try:
                  print("sending " + send_msg)
                  sock.sendto(send_msg, robot_address)
          except:
                  print("FAILED.....Giving up :-(")

def FindColor(imageHSV, lower_col, upper_col, min_area):
    # find the colored regions
    mask=cv2.inRange(imageHSV,lower_col,upper_col)

    # this removes noise by eroding and filling in
    maskOpen=cv2.morphologyEx(mask,cv2.MORPH_OPEN,kernelOpen)
    maskClose=cv2.morphologyEx(maskOpen,cv2.MORPH_CLOSE,kernelClose)
    conts, h = cv2.findContours(maskClose, cv2.RETR_EXTERNAL, cv2.CHAIN_APPROX_NONE)
    # Finding bigest  area and save the contour
    max_area = 0
    for cont in conts:
        area = cv2.contourArea(cont)
        if area > max_area:
            max_area = area
            gmax = max_area
            best_cont = cont
    # identify the middle of the biggest  region
    if conts and max_area > min_area:
        M = cv2.moments(best_cont)
        cx,cy = int(M['m10']/M['m00']),int(M['m01']/M['m00'])
        return best_cont, cx, cy, max_area
    else:
        return 0,-1,-1,-1


lastTime = time.time()
#interval = 0.4
for frame in camera.capture_continuous(rawCapture, format="bgr", use_video_port=True):
    if (time.time()-lastTime) > float(interval):
        lastTime = time.time()
        randval = random.randint(1, 100)
        if (randval < threshold):
            time.sleep(float(interval))
            print("P, I, D, (E), (T) --->", 0, 0, 0, 0, time.time())
            SendToRobot(0,0,0,0,0,0)

    cap_img = frame.array
    full_img = cap_img

    imgHSV = cv2.cvtColor(full_img,cv2.COLOR_BGR2HSV)
    imgHSV_crop = imgHSV[200:280, 0:320]

    key = cv2.waitKey(1) & 0xFF

    best_blackcont, blackcx_incrop, blackcy_incrop, blackarea = FindColor(imgHSV_crop, lower_black, upper_black, 10)

    rawCapture.truncate(0)

    if (blackcx_incrop == -1):
        # if robot not found --> done
        print("P, I, D, (E), (T) --->", 0, 0, 0, 0, time.time())
        SendToRobot(0,0,0,0,0,0)
        continue

    # create a rectangle to represent the line and find the angle of the rectangle on the screen.
    blackbox = cv2.minAreaRect(best_blackcont)
    (x_min, y_min), (w_min, h_min), lineang = blackbox

    blackbox = (x_min, y_min+200), (w_min, h_min), lineang
    drawblackbox = cv2.cv.BoxPoints(blackbox)
    drawblackbox = np.int0(drawblackbox)
    cv2.drawContours(full_img,[drawblackbox],-1,(0,255,0),3)

    # draw line with the estimate of location and angle
    cv2.line(full_img, (int(x_min),int(y_min+200)), (160,40+200), (200,0,200),2)
    cv2.circle(full_img,(int(x_min),int(y_min+200)),3,(200,0,200),-1)
    deltaX = 0.333*(160-x_min)

    cv2.imshow("robotimgPi", full_img)

    P_fix = deltaX
    I_fix = P_fix+0.9*I_fix
    D_fix = P_fix-lastP_fix
    lastP_fix = P_fix
    error = 100*blackarea/5500
    print("P, I, D, (E), (T) --->", P_fix, I_fix, D_fix, error, time.time())

    kP = 1.5
    kI = 0.15
    kD = 4.5
    
    # Compute correction based on angle/position error
    left = int(100 - kP*P_fix - kD*D_fix - kI*I_fix)
    right = int(100 + kP*P_fix + kD*D_fix + kI*I_fix)

    SendToRobot(left,right,error, P_fix, I_fix, D_fix)

    # if the `q` key was pressed, break from the loop
    if key == ord("q"):
    	break

\end{lstlisting}
\newpage
\subsection{Infrastructure Sensor Processing}
\begin{lstlisting}[language=Python]
import cv2
import numpy as np
import os
import socket
import struct
import sys
import time
import math
import random

# set up network socket/addresses
host = '192.168.1.26'
Lport = 4000+int(sys.argv[1])
Rport = 5000
sock = socket.socket(socket.AF_INET, socket.SOCK_DGRAM)
sock.setsockopt(socket.SOL_SOCKET, socket.SO_REUSEADDR, 1)
sock.bind(("", Lport))
print ("Active on port: " + str(Lport))
robot_address = (host, Rport)

# set up camera
camid=sys.argv[1]
cam = cv2.VideoCapture(int(camid))
print("init camera on /dev/video"+sys.argv[1])

#set up opencv variables
kernelOpen=np.ones((5,5))
kernelClose=np.ones((20,20))
font=cv2.FONT_HERSHEY_PLAIN
xdim = 1280
ydim = 720
cropsize = 75
gmax=0
rmax=0
lastP_fix = 0
I_fix=0
colors = []
thiscol = "green"

interval = sys.argv[2]
update = sys.argv[2]
duration = sys.argv[3]
threshold = int(sys.argv[4])

def on_mouse_click (event, x, y, flags, frame):
    global thiscol,lower_green,upper_green,lower_orange,upper_orange,upper_black,lower_black
    if event == cv2.EVENT_LBUTTONUP:
        colors.append(frame[y,x].tolist())
        print(thiscol)
        print(frame[y,x].tolist())
        if thiscol == "green":
            lower_green=np.array([frame[y,x].tolist()[0]-20,frame[y,x].tolist()[1]-30,frame[y,x].tolist()[2]-50])
            upper_green=np.array([frame[y,x].tolist()[0]+20,frame[y,x].tolist()[1]+30,frame[y,x].tolist()[2]+50])
            thiscol = "orange"
        elif thiscol == "orange":
            lower_orange=np.array([frame[y,x].tolist()[0]-20,frame[y,x].tolist()[1]-30,frame[y,x].tolist()[2]-50])
            upper_orange=np.array([frame[y,x].tolist()[0]+20,frame[y,x].tolist()[1]+30,frame[y,x].tolist()[2]+50])
            thiscol = "black"
        elif thiscol == "black":
            lower_black=np.array([frame[y,x].tolist()[0]-20,frame[y,x].tolist()[1]-30,frame[y,x].tolist()[2]-50])
            upper_black=np.array([frame[y,x].tolist()[0]+20,frame[y,x].tolist()[1]+30,frame[y,x].tolist()[2]+50])
            thiscol = "none"

def calibrate():
    capture = cv2.VideoCapture(camid)
    global thiscol
    while thiscol != "none":
        ret, cap_img=cam.read()
        img=cv2.resize(cap_img,(xdim,ydim))
        hsv = cv2.cvtColor(img,cv2.COLOR_BGR2HSV)
        cv2.putText(img, str("CLICK ON " + thiscol), (10, 50), font, 2, (0, 0, 0), 2)
        if colors:
            cv2.putText(img, "LAST: "+str(colors[-1]), (10, 100), font, 2, (0, 0, 0), 2)
        cv2.imshow('frame', img)
        cv2.setMouseCallback('frame', on_mouse_click, hsv)
        if cv2.waitKey(1) & 0xFF == ord('q'):
            break
    capture.release()
    cv2.destroyAllWindows()

def FindColor(imageHSV, lower_col, upper_col, min_area, col):
    # find the colored regions
    mask=cv2.inRange(imageHSV,lower_col,upper_col)
    # this removes noise by eroding and filling in the regions
    maskOpen=cv2.morphologyEx(mask,cv2.MORPH_OPEN,kernelOpen)
    maskClose=cv2.morphologyEx(maskOpen,cv2.MORPH_CLOSE,kernelClose)
    conts, h = cv2.findContours(maskClose, cv2.RETR_EXTERNAL, cv2.CHAIN_APPROX_NONE)
    # Finding bigest area and save the contour
    max_area = 0
    for cont in conts:
        area = cv2.contourArea(cont)
        if area > max_area:
            max_area = area
            gmax = max_area
            best_cont = cont
    # identify the middle of the biggest  region
    if conts and max_area > min_area:
        M = cv2.moments(best_cont)
        cx,cy=int(M['m10']/M['m00']),int(M['m01']/M['m00'])
        return best_cont, cx, cy, max_area
    else:
        return 0,-1,-1,-1

def SendToRobot(left, right, error, P, I, D):
    global sock
    data = str(left)+";"+str(right)+";"+str(error)+";"+str(P)+";"+str(I)+";"+str(D)
    send_msg = str(str(data)).encode()
    try:
          sock.sendto(send_msg, robot_address)
    except Exception as e:
          print("FAIL - RECONNECT.." + str(e.args))
          try:
                  print("sending " + send_msg)
                  sock.sendto(send_msg, robot_address)
          except:
                  print("FAILED.....Giving up :-(")

def ComputeRobotAngle(greencx, greency, orangecx, orangecy):
    # find the angle from the center of green to center
    # of orange this is the angle of the robot in the image
    # I need to special case of 90/-90 due to tan()
    # discontinuity I also need to deal with angles > 90
    # and < 0 to map correctly to a 360 degree circle
    if (greencx-orangecx) == 0:
        if greency > orangecy:
            ang = 90
        else:
            ang = 270
    else:
        Torangecy = ydim - orangecy
        Tgreency = ydim - greency
        ang = 180/np.pi * np.arctan(float(orangecy-greency)/float(orangecx-greencx))
        if greencx > orangecx:
            ang = 180 + ang
        elif ang < 0:
            ang = 360 + ang
        ang = 360-ang
    return ang


calibrate()
lastTime = time.time()

while True:
 try:
    cv2.waitKey(10)
    if (time.time()-lastTime) > float(interval):
        lastTime = time.time()
        if (random.randint(1, 100) < threshold):
            time.sleep(float(interval))
            print("P, I, D, (E), (T) --->", 0, 0, 0, 0, time.time())
            SendToRobot(0,0,0,0,0,0)
            
    try:
        cv2.putText(img, "green: "+str(colors[0]), (10, 50), font, 2, (0, 0, 0), 2)
        cv2.putText(img, "orange: "+str(colors[1]), (10, 80), font, 2, (0, 0, 0), 2)
        dispimg=cv2.resize(img,(720,405))
        cv2.imshow("robotimg"+camid,dispimg)
    except:
        pass

    # grab image, resize, save a copy and convert to HSV
    ret, cap_img=cam.read()
    img=cv2.resize(cap_img,(xdim,ydim))
    imgHSV = cv2.cvtColor(img,cv2.COLOR_BGR2HSV)

    # find largest green region
    best_greencont, greencx, greency, greenarea = FindColor(imgHSV, lower_green, upper_green, 3000, "green")
    if (greencx == -1):
        # if robot not found --> done
        print("ng P, I, D, (E), (T) --->", 0, 0, 0, 0, time.time())
        SendToRobot(0,0,0,0,0,0)
        continue

    cv2.drawContours(img,best_greencont,-1,(0,255,0),3)

    # crop frame to be around robot only
    robotimgHSV = imgHSV[max(greency-200,0):greency+200,max(greencx-300,0):greencx+300]

    # find orange region in this cropped area
    best_orangecont, orangecx_incrop, orangecy_incrop, orangearea = FindColor(robotimgHSV, lower_orange, upper_orange, 3000, "orange")
    if (orangecx_incrop == -1):
        # if robot not found --> done
        print("nr P, I, D, (E), (T) --->", 0, 0, 0, 0, time.time())
        SendToRobot(0,0,0,0,0,0)
        continue

    orangecx = orangecx_incrop+max(greencx-300,0);
    orangecy = orangecy_incrop+max(greency-200,0);
    cv2.drawContours(img, best_orangecont+[max(greencx-300,0),max(greency-200,0)], -1, (0,255,0), 3)

    ang=ComputeRobotAngle(greencx,greency,orangecx,orangecy)

    # draw some robot lines on the screen and display
    cv2.line(img, (greencx,greency), (orangecx,orangecy), (200,0,200),3)
    cv2.putText(img, "robot ang: "+str(ang), (10, 160), font, 2, (0, 0, 0), 2)

    # find a small region in front of the robot and
    # crop that part of the image
    ylen = (greency-orangecy)
    xlen = (greencx-orangecx)
    boxX = orangecx - xlen/2
    boxY = orangecy - ylen/2
    if boxX > (xdim-cropsize):
        Xcropsize = xdim - boxX
    elif boxX < cropsize:
        Xcropsize = boxX
    else:
        Xcropsize = cropsize

    if boxY > (ydim-cropsize):
        Ycropsize = ydim - boxY
    elif boxY < cropsize:
        Ycropsize = boxY
    else:
        Ycropsize = cropsize

    if (Xcropsize > 0) and (Ycropsize > 0):
        lineimgHSV = imgHSV[int(abs(boxY-Ycropsize)):int(abs(boxY+Ycropsize)), int(abs(boxX-Xcropsize)):int(abs(boxX+Xcropsize))]
        # find black region in cropped area
        best_blackcont, blackcx_incrop, blackcy_incrop, blackarea = FindColor(lineimgHSV, lower_black, upper_black, 200, "black")
    else:
        blackcx_incrop = -1

    if (blackcx_incrop == -1):
        # skip if didn't find a line
        print("nb P, I, D, (E), (T) --->", 0, 0, 0, 0, time.time())
        SendToRobot(0,0,0,0,0,0)
        continue

    blackcx = blackcx_incrop+int(abs(boxX-Xcropsize))
    blackcy = blackcy_incrop+int(abs(boxY-Ycropsize))

    # create a rectangle to represent the line and find
    # the angle of the rectangle on the screen.
    blackbox = cv2.minAreaRect(best_blackcont)
    (x_min, y_min), (w_min, h_min), lineang = blackbox

    # draw box on screen
    x_min_real = x_min + int(abs(boxX-Xcropsize))
    y_min_real = y_min + int(abs(boxY-Ycropsize))
    blackbox = (x_min_real, y_min_real), (w_min, h_min), lineang
    drawblackbox = cv2.boxPoints(blackbox)
    drawblackbox = np.int0(drawblackbox)
    cv2.drawContours(img,[drawblackbox],0,(0,255,0),3)
    #cv2.imshow("4",lineimgHSV)

    # Unfortunately, opencv only gives rectangles angles
    # from 0 to -90 so we need to do some guesswork to
    # get the right quadrant for the angle
    if w_min > h_min:
        if (ang > 135):
            lineang = 180 - lineang
        else:
            lineang = -1 * lineang
    else:
        if (ang > 270) or (ang < 45):
            lineang = 270 - lineang
        else:
            lineang = 90 - lineang

    # draw estimate of line location and angle
    x_end = int(x_min_real-50*np.cos(lineang*np.pi/180))
    y_end = int(y_min_real+50*np.sin(lineang*np.pi/180))
    cv2.line(img, (int(x_min_real),int(y_min_real)), (x_end,y_end), (200,0,200),2)
    cv2.circle(img,(int(x_min_real),int(y_min_real)),3,(200,0,200),-1)
    cv2.line(img, (int(x_min_real),int(y_min_real)), (int(boxX),int(boxY)), (200,0,200),2)
    cv2.putText(img, "line ang: "+str(lineang), (10, 190), font, 2, (0, 0, 0), 2)
    #cv2.imshow("5",img)

    # The direction error is the difference in angle of
    # the line and robot essentially the derivative in
    # a PID controller
    D_fix = lineang - ang
    if D_fix < -300:
        D_fix += 360
    elif D_fix > 300:
        D_fix -= 360


    # the line angle guesswork is sometimes off by 180
    # degrees. detect and fix this error here
    if D_fix < -90:
        D_fix += 180
    elif D_fix > 90:
        D_fix -= 180
    cv2.putText(img, "D_fix: "+str(D_fix), (10, 300), font, 2, (0, 0, 0), 2)

    # the position error is an estimate of how far the
    # front center of the robot is from the line. The
    # center of the cropped image
    # (x,y) = (cropsize, cropsize) is the front of the
    # robot. (x_min, y_min) is the center of the line.
    # Draw a line from the front center of the robot
    # to the center of the line. Difference in angle
    # between this line and robot's direction is the
    # position error.
    if (x_min - cropsize) == 0:
        if (ang < 180):
            P_fix = 90 - ang
        else:
            P_fix = 270 - ang
    else:
        temp_angle = 180/np.pi * np.arctan(float(Ycropsize - y_min)/float(x_min - Xcropsize))
        if (temp_angle < 0):
            if (ang > 225):
                temp_angle = 360 + temp_angle
            else:
                temp_angle = 180 + temp_angle
        elif (ang > 135 and ang < 315):
                temp_angle = 180 + temp_angle
        P_fix = temp_angle - ang
    if (P_fix > 180):
        P_fix = P_fix - 360
    elif P_fix < -180:
        P_fix = 360 + P_fix
    # the line angle guesswork is sometimes off by 180
    # degrees. Detect and fix this error here
    if P_fix < -90:
        P_fix += 180
    elif P_fix > 90:
        P_fix -= 180
    cv2.putText(img, "center angle: "+str(temp_angle), (10, 220), font, 2, (0, 0, 0), 2)
    cv2.putText(img, "P_fix: "+str(P_fix), (10, 330), font, 2, (0, 0, 0), 2)

    # Integral controller is just the sum of the P_fix integral P_fix dt ~= sigma P_fix(1)
    # Decay rate of 0.9
    I_fix = P_fix + 0.9*I_fix

    lastP_fix = P_fix

    # print and save correction and current network conditions
    error = 100*blackarea/1300
    print("P, I, D, (E), (T) --->", P_fix, I_fix, D_fix, error, time.time())

    kP = 1.0
    kD = 0.5
    kI = 0.02
        
    # Compute correction based on angle/position error
    left = int(100 - kP*P_fix - kD*D_fix - kI*I_fix)
    right = int(100 + kP*P_fix + kD*D_fix + kI*I_fix)

    # send movement fix to robot
    SendToRobot(left,right,error, P_fix, I_fix, D_fix)

 except:
     pass

\end{lstlisting}

\newpage
\subsection{Vehicle Control Software}

\begin{lstlisting}[language=Python]
#!/usr/bin/python
from Adafruit_MotorHAT import Adafruit_MotorHAT, Adafruit_DCMotor

import socket
import struct
import time
import atexit

HOST = '192.168.1.14' # The remote host
PORT = 5000 # The same port as used by the server
s = socket.socket(socket.AF_INET, socket.SOCK_DGRAM)
s.bind(("",PORT))

address = (HOST, PORT)

cam0=[0,0,0]
cam1=[0,0,0]
picam=[0,0,0]
piout="0,0,0,0,0,0"
cam0out="0,0,0,0,0,0"
cam1out="0,0,0,0,0,0"

try:
   s.sendto("HELLO".encode(),address)
   print("connected to "+HOST)
except Exception as e:
   print("server not available" + str(e.args))
   pass

mh = Adafruit_MotorHAT(addr=0x60)

# recommended for auto-disabling motors on shutdown!
def turnOffMotors():
   mh.getMotor(1).run(Adafruit_MotorHAT.RELEASE)
   mh.getMotor(2).run(Adafruit_MotorHAT.RELEASE)
   mh.getMotor(3).run(Adafruit_MotorHAT.RELEASE)
   mh.getMotor(4).run(Adafruit_MotorHAT.RELEASE)

atexit.register(turnOffMotors)
LMotor = mh.getMotor(3)
RMotor = mh.getMotor(1)

RMotor.run(Adafruit_MotorHAT.FORWARD)
LMotor.run(Adafruit_MotorHAT.BACKWARD)
print(str("time,left,right,piLeft,piRight,piConf,piP,piI,piD,cam0Left,cam0Right,cam0Conf,cam0P,cam0I,cam0D,cam1Left,cam1Right,cam1Conf,cam1P,cam1I,cam1D"))
avgpower = [0,0]
while True:
    # receive messages from server
    data = ''
    try:
        indata = s.recvfrom(1500)
        data,tmp = indata
        ip,port=tmp
        message = data.decode().split(';')
        # divide by 3.0 to make max power only 33%
        left = int(message[0])/3.0 
        message[0] = left
        right = int(message[1])/3.0
        message[1] = right
        confidence = float(message[2])/3.0
        message[2] = confidence
        message[3] = float(message[3])
        message[4] = float(message[4])
        message[5] = float(message[5])

        if ip == "127.0.0.1":
           piout = str(message[0])+","+str(message[1])+","+str(message[2])+","+str(message[3])+","+str(message[4])+","+str(message[5])
           if left > 0 or right > 0:
              picam = message
           else:
              picam = [0,0,0,0,0,0]

        elif ip == "192.168.1.14" and port == 4000:
           cam0out = str(message[0])+","+str(message[1])+","+str(message[2])+","+str(message[3])+","+str(message[4])+","+str(message[5])
           if left > 0 or right > 0:
              cam0 = message
           else:
              cam0 = [0,0,0,0,0,0]
        elif ip == "192.168.1.14" and port == 4001:
           cam1out = str(message[0])+","+str(message[1])+","+str(message[2])+","+str(message[3])+","+str(message[4])+","+str(message[5])
           if left > 0 or right > 0:
              cam1 = message
           else:
              cam1 = [0,0,0,0,0,0]

        # Move motors at power sent from server
        chosen = max([picam[2],cam0[2],cam1[2]])

# Start maximum algorithm
#        if picam[2] == chosen:
#           avgpower=[int(picam[0]),int(picam[1])]
#        if cam0[2] == chosen:
#           avgpower=[int(cam0[0]),int(cam0[1])]
#        if cam1[2] == chosen:
#           avgpower=[int(cam1[0]),int(cam1[1])]
# End maximum algorithm

# Start arithmetic average algorithm
#        camnum = 3
#        if picam[2] == 0:
#           camnum = camnum - 1
#        if cam0[2] == 0:
#           camnum = camnum - 1
#        if cam1[2] == 0:
#           camnum = camnum - 1
#        avgpower = [int((picam[0]+cam0[0]+cam1[0])/(camnum)),int((picam[1]+cam0[1]+cam1[1])/(camnum))]
# End arithmetic average algorithm

# Start weighted average algorithm
        avgpower = [int((picam[2]*picam[0]+cam0[2]*cam0[0]+cam1[2]*cam1[0])/(picam[2]+cam0[2]+cam1[2])),int((picam[2]*picam[1]+cam0[2]*cam0[1]+cam1[2]*cam1[1])/(picam[2]+cam0[2]+cam1[2]))] 
# End weighted average algorithm

        print(str(time.time())+","+str(avgpower[0])+","+str(avgpower[1]) + "," +str(piout)+","+str(cam0out)+","+str(cam1out))
        LMotor.setSpeed(avgpower[0])
        RMotor.setSpeed(avgpower[1])
    except Exception as e:
        print(str(time.time())+","+str(avgpower[0])+","+str(avgpower[1]) + "," +str(piout)+","+str(cam0out)+","+str(cam1out)+",-1")
        try:
           s.close()
           s=socket.socket(socket.AF_INET,socket.SOCK_DGRAM)
           s.bind(("",PORT))
        except:
           pass


\end{lstlisting}

\end{document}